\documentclass[10pt,twocolumn,letterpaper]{article}

\usepackage[pagenumbers]{cvpr} %

\usepackage{amsmath}
\usepackage{amssymb}
\usepackage{amsfonts}
\usepackage{mathabx}
\usepackage[table]{xcolor}   %

\definecolor{tabgreen1}{HTML}{71AD8B} %
\definecolor{tabgreen2}{HTML}{B6D6B5} %
\definecolor{tabgreen3}{HTML}{E5EFE0} %

\colorlet{firstcolor}{tabgreen1}
\colorlet{secondcolor}{tabgreen2}
\colorlet{thirdcolor}{tabgreen3}

\newcommand{\best}{\cellcolor{firstcolor}}
\newcommand{\better}{\cellcolor{secondcolor}}
\newcommand{\good}{\cellcolor{thirdcolor}}

\newcommand{\dejaview}{D\'ej\`aView\xspace}

\DeclareMathOperator*{\argmin}{arg\,min}

\usepackage{pifont}
\definecolor{cmarkgreen}{HTML}{2E8B57}
\definecolor{xmarkred}{HTML}{C0392B}

\newcommand{\cmark}{\textcolor{cmarkgreen}{\ding{51}}}
\newcommand{\xmark}{\textcolor{xmarkred}{\ding{55}}}

\definecolor{cvprblue}{rgb}{0.21,0.49,0.74}
\usepackage[pagebackref,breaklinks,colorlinks,citecolor=cvprblue,hypertexnames=false]{hyperref}

\usepackage{colortbl}
\usepackage{pifont}
\usepackage{cuted}      %
\usepackage{afterpage}  %
\newcommand{\RN}[1]{\uppercase\expandafter{\romannumeral #1}}

\newcommand{\ours}{\texorpdfstring{VGGT-$\mu$}{VGGT-mu}}

\def\confName{CVPR}
\def\confYear{2027}

\title{\ours: Lightweight Relative Pose Regression at Scale}

\author{
Ludvig Dillén$^{1, 2}$\thanks{Equal contribution. Listing order is random.} \hspace{2em} David Nordström$^{3}$\textsuperscript{\thefootnote} \hspace{2em} Xinyue Zhang$^{4}$ \hspace{2em} Yaroslava Lochman$^{3}$ \\
Janne Heikkilä$^{4}$ \hspace{2em} Fredrik Kahl$^{3}$ \hspace{2em} Viktor Larsson$^{1}$ \\
{\small $^1$Lund University \hspace{2em} $^2$KTH Royal Institute of Technology \hspace{2em} $^3$Chalmers University of Technology \hspace{2em} $^4$University of Oulu} \\
}

\begin{document}
\raggedbottom
\maketitle

\begin{strip}
\vspace{-20pt}
\centering
\includegraphics[width=\textwidth]{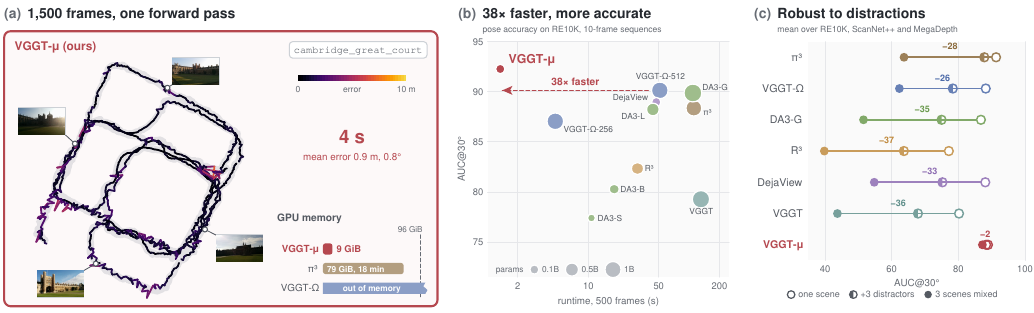}
\vspace{-6pt}
\captionsetup{type=figure,font=small}
\caption{\textbf{Camera poses for 1,500 frames in one forward pass.}
\textbf{(a)}~VGGT-$\mu$ estimates camera poses for 1,500 frames in 4 seconds using 9\,GiB (0.9\,m and 0.8$^\circ$ mean errors), while VGGT-$\Omega_{512}$ runs out of memory and $\pi^3$ needs 18\,min and 79\,GiB (1.8\,m and 2.2$^\circ$).
\textbf{(b)}~It achieves the best RE10K AUC@30$^\circ$ and is 38$\times$ faster than the next best method, VGGT-$\Omega_{512}$.
\textbf{(c)}~VGGT-$\mu$ is robust to distractor images and mixed-scene inputs. Mixing three scenes reduces baseline AUC@30$^\circ$ by 26--37 percentage points, compared with only 2 for VGGT-$\mu$.
}
\label{fig:teaser}
\vspace{-8pt}
\end{strip}

\afterpage{\flushbottom}

\begin{abstract}
Estimating camera poses from a set of images is a core problem in 3D vision. Feed-forward reconstruction models, such as VGGT, tackle this problem by jointly predicting camera poses and depth maps in a single forward pass. While powerful, these models' runtime and memory requirements grow rapidly with the number of input images. In practice, many applications, such as visual localization and odometry, require only camera poses. Thus, we propose \ours, a lightweight recurrent model that replaces feed-forward reconstruction with feed-forward pose regression, scaling to thousands of images thanks to low-resolution input, low parameter count, and no depth prediction. With only 120M parameters, \ours~rivals its bigger brother, VGGT-$\Omega$, on pose benchmarks. Relying on relative pose regression, \ours~handles disconnected view graphs while providing metric poses and information matrices for anisotropic pose averaging. 
For example, \ours~reconstructs 1,500 frames in 4 seconds using 9\,GiB of VRAM, whereas VGGT-$\Omega$ runs out of memory and $\pi^3$ is less accurate despite requiring 18 minutes and 79\,GiB. Code and model weights at \href{https://github.com/davnords/vggt-mu}{https://github.com/davnords/vggt-mu}.
\end{abstract}
\section{Introduction} \label{sec:intro}

Relative pose estimation has applications in autonomous driving, augmented reality, robot navigation, and visual localization. 
Traditionally, relative poses are obtained by finding sparse matches~\cite{lowe2004distinctive} followed by robust geometric estimation (e.g.~RANSAC)~\cite{Hartley_Zisserman_2004}. 
In recent years, this pipeline has been augmented by more and more learning-based components.
Beginning with DUSt3R~\cite{wang2024dust3r}, feed-forward reconstruction methods now directly estimate the 3D scene geometry using a neural network, 
and scaling these models has culminated in the current flagship, VGGT-$\Omega$~\cite{wang2026vggtomega}. While the model yields accurate results, using it to estimate multi-view relative poses is often excessive and scales poorly in runtime and memory for long sequences. 
This motivates the need for smaller, purpose-built relative pose estimators that retain the robust priors of VGGT-$\Omega$.

There is precedent for converting feed-forward reconstruction models into relative pose estimators. In the era of pairwise models, Reloc3r~\cite{reloc3r:2025} adapted DUSt3R to relative pose estimation. In a similar spirit, we adapt VGGT-$\Omega$ specifically for relative pose estimation from image collections. To enable lightweight inference, we address two distinct bottlenecks that emerge at different sequence lengths. For short sequences, VRAM usage is dominated by model weights, which we reduce through weight sharing across transformer blocks~\cite{burzio2026dejaview} and dropping the depth head. For longer sequences, global attention becomes the primary computational bottleneck, with runtime scaling as $\mathcal{O}(N^2R^4)$ for $N$ images at resolution $R$. We therefore design \ours~to operate at low resolution, which we show is sufficient for accurate pose estimation. 

Efficiency alone, however, is insufficient for real-world image collections, which are often unordered and may contain multiple disjoint components. %
Existing feed-forward models typically assume a single scene and, as illustrated in \cref{fig:teaser}~(c), struggle with \textit{distractors} and disconnected view graphs, producing a joint coordinate frame even for unrelated inputs.
In contrast, \ours~directly predicts pairwise relative poses and confidences and is trained on disconnected view graphs to distinguish unrelated views while accurately estimating matching pairs, making it well suited for rapid view-graph initialization in long sequences. \ours~further predicts metric pose scale, $6\times6$ pairwise information matrices enabling uncertainty-aware pose averaging~\cite{olsson2025certifiably}, and supports conditioning on known poses and intrinsics. Together, these capabilities yield \ours\footnote{A micro ($\mu$) version of VGGT-$\Omega$.}, a fast and robust relative pose estimator with a favorable accuracy--efficiency trade-off, as shown in \cref{fig:teaser}.

\noindent\textbf{The main contributions are as follows:}
\begin{enumerate}[label=(\roman*), topsep=0pt]
\item \textbf{A lightweight and accurate relative pose estimator.} We adapt VGGT-$\Omega$ into a substantially more efficient model through weight sharing and low-resolution inference, while retaining competitive accuracy across relative pose estimation and downstream tasks.

\item \textbf{Robust and uncertainty-aware pose estimation.} \ours~is trained to handle unordered image collections containing disconnected view graphs and distractor images while jointly predicting pose information matrices that provide  calibrated uncertainty.

\item \textbf{Versatile camera-pose estimation.} \ours~supports metric pose prediction and conditioning on known 
camera poses and intrinsics, enabling flexible integration into visual localization, Structure-from-Motion (SfM), and visual odometry (VO) pipelines.
\end{enumerate}

\begin{figure*}[t]
\centering
\includegraphics[width=0.99\linewidth]{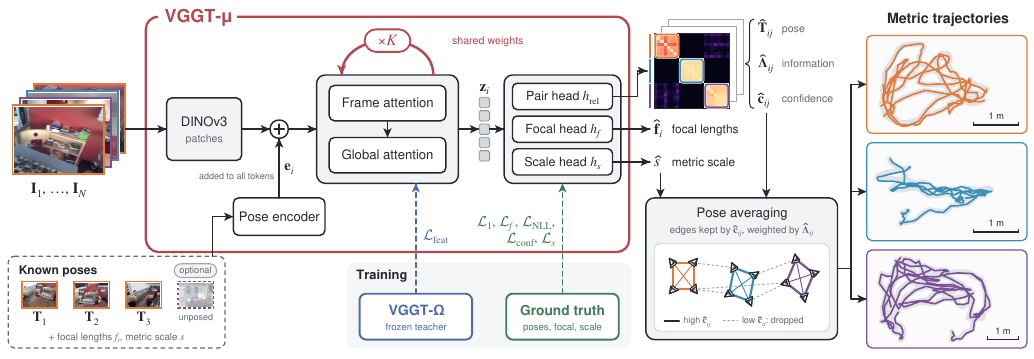}
\caption{\textbf{\ours} takes as input a set of images, possibly from different scenes (3 in this example), and alternates between frame-wise and global attention. Pairs of learnable camera tokens are concatenated and input into an MLP that predicts pairwise confidences, information matrices, and poses. To stabilize training, we use feature distillation from VGGT-$\Omega$. Anisotropic pose averaging yields absolute poses.}
\label{fig:method}
\end{figure*}

\section{Related Work} \label{sec:related}

\paragraph{3D vision pipelines.}
Classical SfM~\cite{schoenberger2016sfm,pan2024glomap} and SLAM~\cite{mur2015orb,ORBSLAM3_TRO,teed2021droid} have adopted learned features and matchers~\cite{SuperPoint:2018,sarlin2020superglue,zhao2022alike,Zhao2023ALIKED,nordstrom2026lomalocalfeaturematching}, as well as feed-forward reconstruction priors~\cite{brynte2024learning,pan2026gluemap,durso2026epo,Taher_2026_LatentBA,duisterhof2025mast3r,maggio2026vggt,mandal2026vggt,maggio2026vggt2}. GlueMap~\cite{pan2026gluemap}, for example, reconstructs image clusters using a feed-forward model and integrates the estimates within GLOMAP~\cite{pan2024glomap}. In contrast, \ours~directly predicts relative poses for the full image collection in a single forward pass, which we use to initialize a global SfM pipeline. Feed-forward reconstruction also supports SLAM systems such as VGGT-SLAM 2.0~\cite{maggio2026vggt2}, but its GPU memory requirements and submap-wise pose updates can limit applications with tight memory and latency budgets. We evaluate \ours~as a lightweight alternative for VO. Structureless visual localization estimates a query pose from reference images with known poses, without an explicit 3D map~\cite{panek2026guide,dillen2026sparse}. We thus evaluate \ours~by conditioning on reference poses and, when available, camera intrinsics.

\paragraph{Feed-forward reconstruction and pose regression.} Feed-forward reconstruction models estimate 3D geometry directly from images. DUSt3R~\cite{wang2024dust3r} and MASt3R~\cite{leroy2024grounding} introduced pairwise reconstruction, which has subsequently been extended to multi-view, pose-conditioned, and metric reconstruction~\cite{wang2025vggt,wang2026vggtomega,jang2025pow3r,keetha2026mapanything,depthanything3,wang2025pi3}.
Several inference speedups enable long-sequence reconstruction~\cite{shen2026fastvggt,shu2026litevggtboostingvanillavggt,wang2026flashvggtefficientscalablevisual,wang2026blocksparseglobalattentionefficient} and are compatible with our approach (\cref{fig:pareto-long}). Reloc3r~\cite{reloc3r:2025} directly regresses two-view relative poses, while $\pi^3$~\cite{wang2025pi3} uses relative pose supervision, and $R^3$~\cite{xu2026r33dreconstructionrelative} predicts multi-view relative poses alongside dense geometry. We similarly regress relative poses, but specialize in lightweight multi-view pose estimation without dense geometry prediction. \dejaview~\cite{burzio2026dejaview} introduces looping layers, which we adopt for parameter efficiency.
\citet{paul2026viewssceneevaluatingmultiview} studied robustness to mixed-scene inputs, and \citet{ziliotto2026covggt} used VGGT for co-visibility prediction. Concurrently to ours, Graph3R~\cite{yu2026graph3runifyingviewgraph} uses a pairwise co-visibility head to group disconnected view graphs, with a $\pi^3$ backbone and no joint reconstruction supervision, whereas our lightweight model jointly learns relative poses and edge reliability.

\paragraph{Uncertainty in 3D vision.} Most feed-forward reconstruction methods predict per-pixel confidences~\cite{wang2024dust3r,wang2025vggt}, as do many image matching methods~\cite{edstedt2023dkm,edstedt2024roma,zhang2025ufm}. $R^3$~\cite{xu2026r33dreconstructionrelative} goes beyond this by predicting scalar confidences for pose edges, while RoMa v2~\cite{edstedt2026romav2} predicts $2\times2$ information matrices for image warps. \ours~extends this idea to pose edges by predicting full $6\times6$ information matrices, enabling uncertainty-aware pose averaging~\cite{olsson2025certifiably,lochman2025making}.

\section{Method}

In this section, we first introduce the problem setup and relative pose formulation (\cref{subsec:problem,subsec:relative-pose}), followed by the camera loss supervision and pose averaging (\cref{subsec:loss,sec:pose-avr}), and our approach to multiple scenes (\cref{subsec:disjoint}). We then describe the architecture, distillation strategy, and conditioning (\cref{subsec:architecture,subsec:distillation,subsec:conditioning}). An overview of our method is given in \cref{fig:method} and a qualitative example in \cref{fig:qualitative}. 

\subsection{Problem Setup}\label{subsec:problem}

Given the input images $\mathcal{I}=\{\mathbf{I}_i\}_{i=1}^N$, where $\mathbf{I}_i\in \mathbb{R}^{H\times W\times3}$, our goal is to recover the world-to-camera poses $\mathbf{T}_i = (\mathbf{R}_i, \mathbf{t}_i)$ and intrinsics. Here, $\mathbf{R}_i \in SO(3)$ and $\mathbf{t}_i \in \mathbb{R}^3$ denote the camera rotation and translation, respectively. Similar to earlier work~\cite{wang2024dust3r,wang2025vggt}, we assume zero skew and that the principal point is at the image center, and thus only estimate the horizontal and vertical focal lengths $\mathbf{f}_i=(f_i^x, f_i^y)$. %

\subsection{Relative Pose Formulation}\label{subsec:relative-pose}
In feed-forward reconstruction, camera poses $\mathbf{T}_i$ are typically predicted directly in a global coordinate frame. Instead, following \citet{xu2026r33dreconstructionrelative}, we predict pairwise relative poses (see \cref{suppl:rel-vs-abs-pose} for an analysis of their consistency across different paths between the same cameras). We extend their uncertainty representation of the pairwise estimates from mere confidences to $6\times6$ information matrices (inverse covariances) describing the full coupling of the $6$-DoF relative poses we predict. Our neural network consists of a backbone $g_\phi$ and three heads, $h_f$, $h_\mathrm{rel}$, and $h_s$. The transformer backbone $g_\phi$ processes images as
\begin{equation}
      \{ \mathbf{z}_i\}_{i=1}^N=g_\phi(\mathcal{I}), 
\end{equation}
where $\mathbf{z}_i\in \mathbb{R}^D$ are the latent camera tokens and $D$ is the feature dimension. 
Each camera token $\mathbf{z}_i$ is decoded into focal lengths
$\hat{\mathbf{f}}_i = h_f(\mathbf{z}_i)$, while a separate head
predicts the metric scale from the average-pooled camera tokens, $\hat{s} = h_s\bigl(\frac{1}{N}\sum_{i=1}^{N}\mathbf{z}_i\bigr)$.
Pairs of camera tokens are concatenated and passed to the MLP
$h_\mathrm{rel}$ to predict relative poses ($i\to j$),
confidences, and information matrices:
\begin{equation}
    \bigl(\hat{\mathbf{T}}_{ij},\,
          \hat{c}_{ij},\,
          \hat{\boldsymbol{\Lambda}}_{ij}\bigr)
    = h_\mathrm{rel}([\mathbf{z}_i; \mathbf{z}_j]),
\end{equation}
where $1 \le i,j \le N$ and $i \neq j$.
Pairwise estimates are combined into absolute poses by pose averaging (\cref{sec:pose-avr}). %

\subsection{Camera Loss Supervision}\label{subsec:loss}
We detail the camera loss terms, including the information matrices, confidences, and metric scale. \Cref{subsec:distillation} introduces an additional loss term for VGGT-$\Omega$ feature distillation. The full loss combines these into $\mathcal{L} = \mathcal{L}_{\mathrm{cam}}+\mathcal{L}_\mathrm{feat}$. Our camera loss averages four terms over all edges $\mathcal{E}=\{(i,j) : i\neq j\}$ in the same scene (with confidence supervised over all pairs) plus a scale loss:
\begin{equation}
\begin{aligned}
    &\mathcal{L}_{\mathrm{cam}}
    = \frac{\lambda_{\mathrm{cam}}}{|\mathcal{E}|}\sum_{(i,j)\in\mathcal{E}}
    \Bigl(\lambda_{1}\,\mathcal{L}_{1}(i,j)+ \lambda_{f}\,\mathcal{L}_{f}(j) +
      \\ &\lambda_{\mathrm{NLL}}\,\mathbf{1}_{(i,j)\in\mathcal{C}}\,\mathcal{L}_{\mathrm{NLL}}(i,j)+\lambda_{\mathrm{c}}\,\mathcal{L}_{\mathrm{conf}}(i,j)
    \Bigr) + \lambda_s\mathcal{L}_s,
\end{aligned}
\end{equation}
with $\lambda_{\mathrm{cam}}=\lambda_{s}=0.05$, $\lambda_1=\lambda_{\mathrm{NLL}}=\lambda_{\mathrm{c}}=1$, and $\lambda_{f}=0.5$. We detail each term below.

\paragraph{Pose residual.}
Ground-truth (GT, denoted by $^*$) relative poses $\mathbf{T}^{*}_{ij}$ are computed from the GT trajectory, expressed relative to the first camera and normalized such that the camera centers are on average at unit distance. We measure the error of a prediction in the tangent space of $\mathrm{SE}(3)$:
\begin{equation}\label{eq:pose_residual}
    \mathbf{r}_{ij}^*
    = \mathrm{Log}\!\left(\hat{\mathbf{T}}_{ij}^{-1}\,\mathbf{T}^{*}_{ij}\right)
    = \begin{bmatrix}\boldsymbol{\omega}_{ij}^*\\\boldsymbol{\rho}_{ij}^*\end{bmatrix}\in\mathbb{R}^6,
\end{equation}
where $\boldsymbol{\omega}$ and $\boldsymbol{\rho}$ are the rotation and translation parts. We also use two angular errors, which do not depend on scale. The first is the rotation error $\theta^{\mathrm{R}}_{ij}=\|\boldsymbol{\omega}_{ij}^*\|_2$. The second is the translation-direction error $\theta^{\mathrm{t}}_{ij}=\angle\bigl(\hat{\mathbf{t}}_{ij},\mathbf{t}^{*}_{ij}\bigr)$.

\begin{figure*}[t]
\centering
\includegraphics[width=0.99\linewidth]{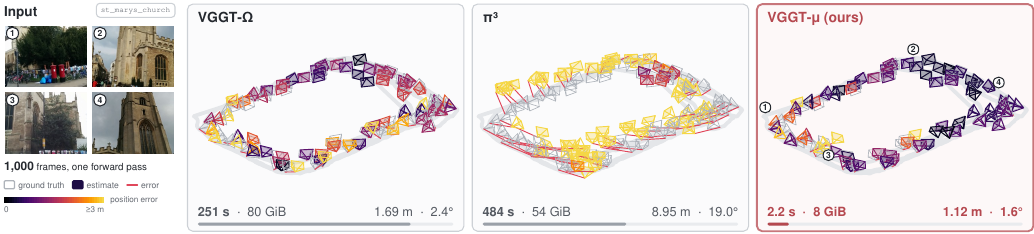}
\caption{\textbf{Qualitative results.} \ours~estimates poses for 1,000 frames in two seconds while being more accurate than $\pi^3$ and VGGT-$\Omega$.}
\label{fig:qualitative}
\end{figure*}

\paragraph{Pose regression.}
All edges are supervised with an $L_1$ loss on the residual,
$
    \mathcal{L}_{1}(i,j) = \|\mathbf{r}_{ij}^*\|_1.
$
This is important since the likelihood term below only covers \emph{close} edges. $\mathcal{L}_1$ is therefore the only pose signal for all other edges.

\paragraph{Focal length.}
The per-frame intrinsics from $h_f$ are supervised with an $L_1$ loss,
$\mathcal{L}_{f}(j)=\|\hat{\mathbf{f}}_j-\mathbf{f}^{*}_j\|_1$.

\begin{figure}[t]
\centering
\includegraphics[width=0.99\linewidth]{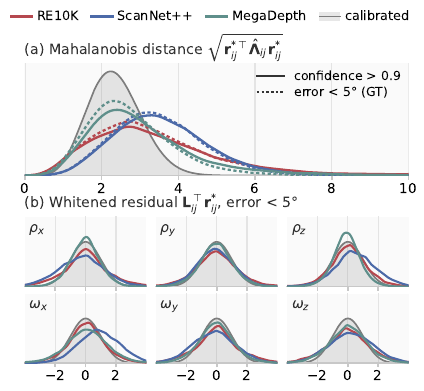}
\caption{\textbf{Uncertainty calibration.} \textbf{(a)} Mahalanobis distance of relative pose residuals $\mathbf{r}_{ij}^*$ under the predicted information matrix $\hat{\boldsymbol{\Lambda}}_{ij}=\mathbf{L}_{ij}\mathbf{L}_{ij}^{\top}$, against the $\chi_6$ distribution expected under calibration.
\textbf{(b)} Whitened residual components $\mathbf{L}_{ij}^{\top}\mathbf{r}_{ij}^*$, against $\mathcal{N}(0,1)$.}
\label{fig:uncertainty}
\end{figure}

\paragraph{Information matrices.}
The head $h_\mathrm{rel}$ outputs 21 parameters that fill the lower-triangular Cholesky factor $\mathbf{L}_{ij}\in\mathbb{R}^{6\times6}$ of the information matrix. Its diagonal passes through a softplus plus a small constant, so that
$
    \hat{\boldsymbol{\Lambda}}_{ij}=\mathbf{L}_{ij}\mathbf{L}_{ij}^{\top}
$
is positive definite by construction. We train $\hat{\boldsymbol{\Lambda}}$ by minimizing the Gaussian negative log-likelihood (NLL) of the residual
\begin{equation}
    \mathcal{L}_{\mathrm{NLL}}(i,j)
    = \left(\mathbf{r}_{ij}^*\right)^{\top}\hat{\boldsymbol{\Lambda}}_{ij}\mathbf{r}_{ij}^*
    - \log\det\hat{\boldsymbol{\Lambda}}_{ij}.
\end{equation}
The Gaussian approximation is only accurate when the error is small, and large residuals also make the NLL unstable early in training. We therefore apply this term only to \emph{close} edges, $\mathcal{C}=\{(i,j)\in\mathcal{E} : \theta^{\mathrm{R}}_{ij}<\tau,\ \theta^{\mathrm{t}}_{ij}<\tau\}$ with $\tau=5^\circ$. On these edges, the gradient of the NLL flows into both $\hat{\boldsymbol{\Lambda}}$ and the predicted pose. We verify the quality of $\hat{\boldsymbol{\Lambda}}$ in \cref{fig:uncertainty}. In (a), Mahalanobis distances roughly follow the expected $\chi_6$ distribution but are slightly larger, indicating mildly overconfident information matrices. In (b), whitened residuals are approximately normal for most pose components and datasets, except $\rho_x$, $\rho_z$, and $\omega_x$ in ScanNet++.

\paragraph{Confidence.}
$\hat{\boldsymbol{\Lambda}}$ only describes the local error on edges. We therefore also predict a scalar confidence $\hat c_{ij}=\sigma(\cdot)\in(0,1)$ for overall edge reliability. Its target and loss are
\begin{equation}
    c^{*}_{ij} = 1-\frac{\theta^{\mathrm{R}}_{ij}+\theta^{\mathrm{t}}_{ij}}{2\pi},  \quad \mathcal{L}_{\mathrm{conf}}(i,j) = \bigl(\hat c_{ij}-c^{*}_{ij}\bigr)^2
\end{equation}

\paragraph{Metric scale.}\label{subsec:metric}
Following~\cite{wang2025moge2}, we supervise $\hat{s}$ with
\begin{equation}
    \mathcal{L}_s=\bigl|\log\hat{s}-\log s^*\bigr|,
\end{equation}
where $s^*$ is the GT scale (see \cref{suppl:gt-scale}, Suppl.). Similar to MASt3R~\cite{leroy2024grounding}, we apply this loss only to metric-scale data. Metric datasets are marked in \cref{tab:dataset_mix} (Suppl.).

\begin{figure*}[t]
\centering
\includegraphics[width=0.99\linewidth]{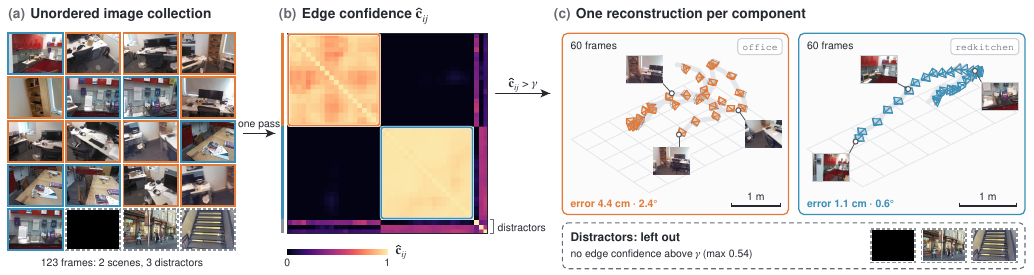}
\caption{\textbf{Robustness via edge confidences.} VGGT-$\mu$ predicts pairwise edge confidences $\hat{c}_{ij}$ for two unordered 7-Scenes~\cite{shotton2013scene} sequences and three \textit{distractors}. Thresholding at $\gamma$ separates scenes and rejects distractors. Here, $\gamma\in[0.54,0.9]$ achieves perfect separation.}
\label{fig:confmat}
\end{figure*}

\subsection{Pose Averaging}\label{sec:pose-avr}
While the network predicts pairwise relative poses, many downstream applications require every camera expressed in a single coordinate frame. We obtain these absolute poses through anisotropic $\mathrm{SE}(3)$-averaging, with the help of the predicted information matrices. Let $\mathbf{R}_{ij}=\mathbf{R}_j\mathbf{R}_i^\top$ denote the relative rotation implied by the absolute poses. We use the same rotational residual as in \cref{eq:pose_residual},
\begin{align}
\boldsymbol{\omega}_{ij}
&= \mathrm{Log}\!\left(
\hat{\mathbf{R}}_{ij}^{\top}\mathbf{R}_{ij}
\right),
\end{align}
where $\hat{\mathbf{R}}_{ij}$ is the predicted relative rotation, and optimize for the absolute rotations using anisotropic averaging%
\begin{align}
\{\tilde{\mathbf{R}}_k\}_{k=1}^{N}
&=
\argmin_{\{\mathbf{R}_k\}_{k=1}^N\subseteq SO(3)}
\sum_{(i,j)\in\mathcal{E}}
\boldsymbol{\omega}_{ij}^\top \boldsymbol{\Lambda}^{\omega}_{ij} \boldsymbol{\omega}_{ij}.
\end{align}
Here, $\boldsymbol{\Lambda}^{\omega}_{ij}$ are obtained from $\hat{\boldsymbol{\Lambda}}_{ij}$ by marginalizing out translations. We use the objective of \citet{olsson2025certifiably}, which approximates the terms $\boldsymbol{\omega}_{ij}^\top \boldsymbol{\Lambda}^{\omega}_{ij} \boldsymbol{\omega}_{ij}$ with bilinear terms in $\mathbf{R}_i$ and $\mathbf{R}_j$,
and we locally optimize it with the solver~\cite{lochman2025making}.
For translation averaging, we use the residual:
\begin{align}
\boldsymbol{\epsilon}_{ij} &= (\mathbf{t}_j-\hat{\mathbf{R}}_{ij}\mathbf{t}_i)
-\hat{\mathbf{t}}_{ij},
\end{align}
 where $\hat{\mathbf{t}}_{ij}$ is the predicted relative translation.
We then use the sub-blocks $\boldsymbol{\Lambda}^{\mathrm{t}}_{ij}$ of the predicted information matrices $\hat{\boldsymbol{\Lambda}}_{ij}$ to optimize the absolute translations
\begin{align}
\{\tilde{\mathbf{t}}_k\}_{k=1}^{N}
&=
\argmin_{\{\mathbf{t}_k\}_{k=1}^N\subseteq \mathbb{R}^3}
\sum_{(i,j)\in\mathcal{E}}
\boldsymbol{\epsilon}_{ij}^\top \boldsymbol{\Lambda}^{\mathrm{t}}_{ij} \boldsymbol{\epsilon}_{ij},
\end{align}
which is a weighted least-squares problem. Since the rotations are fixed at this stage, we use the translation block $\boldsymbol{\Lambda}^{\mathrm{t}}_{ij}$ directly rather than marginalizing as for rotations. It is the information of the translation conditioned on the rotation.

\subsection{Robustness to Distractors and Multiple Scenes}\label{subsec:disjoint}
  \label{sec:mixing}

Enabled by our pairwise relative pose formulation, we train on unordered image collections containing multiple disjoint scenes. Specifically, when constructing sequences of length $N$, we sample pairs from independent scenes with probability $p=0.3$. Pose and uncertainty losses are applied only to same-scene pairs, while cross-scene pairs are assigned zero confidence. This yields a block-diagonal confidence matrix with one block per scene (\cref{fig:confmat}). At inference, we use $1-\hat{c}_{ij}$ as a pairwise distance, apply average-linkage clustering, and cut at a threshold $\gamma$ to separate scenes.

\subsection{Architecture}\label{subsec:architecture}

\ours~follows VGGT's alternating frame-wise and global attention design~\cite{wang2025vggt,wang2026vggtomega}, but replaces its stack of distinct layers with a recurrent block with shared weights, inspired by \dejaview~\cite{burzio2026dejaview}. The block is applied $K$ times and conditioned on its iteration via learned gates, allowing the shared block to specialize across iterations. During training, $K\in[8,16]$ is sampled per batch by rescaling and rounding a $\mathrm{Beta}(2,1)$ sample, while we use $K=12$ at inference, making $K$ a test-time compute knob. We use a ViT-B encoder initialized from DINOv3~\cite{simeoni2025dinov3} and $D=384$, yielding a 120M-parameter model (see \cref{fig:param-breakdown}, Suppl.). %

\subsection{Distillation Protocol}\label{subsec:distillation}

The camera loss in \cref{subsec:loss} is combined with feature distillation from a frozen VGGT-$\Omega_{256}$ teacher.
For each distillation level $m\in\{1,\ldots,4\}$, corresponding to teacher layer $\ell_m\in\{4,11,17,23\}$, let $\mathbf{v}_{m,t}\in\mathbb{R}^{2D_{\mathrm{tea}}}$ denote the teacher's feature of token $t$, obtained by concatenating the frame- and global-attention outputs. Since the student reuses a single block for $K$ iterations, we extract the corresponding student feature $\mathbf{u}_{m,t}\in\mathbb{R}^{2D_{\mathrm{stu}}}$ at iteration $\operatorname{round}(\alpha_m(K-1))$, counting from 0, with $\alpha_m\in\{\frac14,\frac12,\frac34,1\}$. We minimize
\begin{equation}
\mathcal{L}_{\mathrm{feat}}
=\frac{1}{|\mathcal{T}|}\sum_{m=1}^{4}\sum_{t\in\mathcal{T}}
\mathcal{H}\bigl(\tilde{\mathbf{u}}_{m,t}-\tilde{\mathbf{v}}_{m,t}\bigr),
\end{equation}
where $\mathcal{T}$ contains all tokens, $\tilde{\mathbf{u}}_{m,t}=\operatorname{LN}(\mathbf{P}_m\mathbf{u}_{m,t})$ and $\tilde{\mathbf{v}}_{m,t}=\operatorname{LN}(\mathbf{v}_{m,t})$ are the LayerNorm-normalized student and teacher features, $\mathbf{P}_m$ is a linear projector discarded after training, and $\mathcal{H}$ is the Huber loss.  %

\subsection{Conditioning on Known Quantities}\label{subsec:conditioning}

If some poses $\{\mathbf{T}_i\}_{i\in\mathcal{K}}$ or focal lengths $\{\mathbf{f}_i\}_{i\in\mathcal{F}}$ are known (\eg, from a rig, odometry, or a mapped scene), we can condition on them.
Known poses (at least two) are expressed relative to the first known frame $r$ and their translations divided by $\bar{s}=\frac{1}{|\mathcal{K}|}\sum_{i\in\mathcal{K}}\|\mathbf{t}_{ri}\|$, giving unit-free poses $\bar{\mathbf{T}}_i$.
If the known poses are metric, we can additionally provide their scale $\bar{s}$.
Each image then receives the embedding
\begin{equation}\label{eq:conditioning}
\mathbf{e}_i
=\delta^{\mathbf{T}}_i\,\phi_{\mathbf{T}}(\bar{\mathbf{T}}_i)
+\delta^{\mathrm{f}}_i\,\phi_{\mathrm{f}}(\mathbf{f}_i)
+\delta^{\mathrm{s}}_i\,\phi_{\mathrm{s}}(\log\bar{s}),
\end{equation}
where $\delta_i\in\{0,1\}$ indicates what is known and each $\phi$ is a small MLP. The embedding $\mathbf{e}_i$ is added to all tokens of $\mathbf{I}_i$ before $g_\phi$. At inference, edges between known frames are set to their given values with $\hat{c}_{ij}=1$ before pose averaging, and the result is rescaled by $\bar{s}$.

\section{Results} \label{sec:results}

In this section, we first show runtime and memory analyses (\cref{subsec:results:runtime}), followed by multi-view relative pose estimation (\cref{subsec:results:mvrelpose}). We then evaluate \ours~in three downstream settings: visual localization, SfM, and VO (\cref{subsec:results:visloc,subsec:results:sfm,subsec:results:vo}). Lastly, we perform scene clustering (\cref{subsec:results:mvrelpose-mixing}) and ablate our design choices (\cref{subsec:results:ablations}). Generally, the \colorbox{firstcolor}{best}, \colorbox{secondcolor}{second}, and \colorbox{thirdcolor}{third}-ranked results are highlighted.

\paragraph{Implementation details.} \ours~is trained for 1M steps using  AdamW~\cite{adamw:2019} with a cosine learning rate schedule peaking at $2\times10^{-4}$ on the data in \cref{tab:dataset_mix} (Suppl.). We sample sequence lengths of 2–24 and pack sequences up to 96 frames per GPU. Training takes roughly nine days on 16 GH200 GPUs at $256\times256$ resolution.

\paragraph{Baselines.} We compare to state-of-the-art feed-forward reconstruction models, VGGT-$\Omega$~\cite{wang2026vggtomega}, VGGT~\cite{wang2025vggt}, $\pi^3$~\cite{wang2025pi3}, $R^3$~\cite{xu2026r33dreconstructionrelative}, Depth Anything 3 (DA3)~\cite{depthanything3}, and \dejaview~\cite{burzio2026dejaview}. %

\subsection{Runtime and Memory}\label{subsec:results:runtime}

\begin{figure}[t]
\centering
\includegraphics[width=0.99\linewidth]{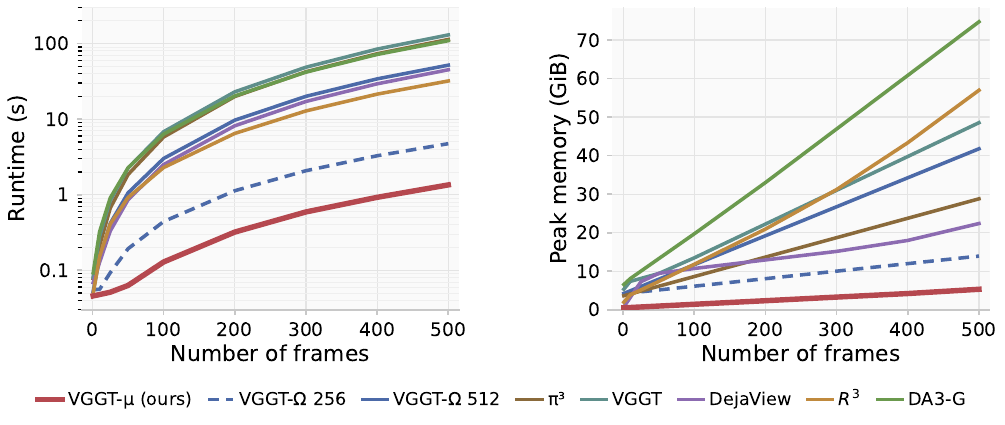}
\caption{\textbf{Memory and speed.} We run inference on an NVIDIA GH200 GPU in half precision with FlashAttention-3~\cite{shah2024flashattention3fastaccurateattention} enabled.}
\label{fig:runtime}
\end{figure}

In \cref{fig:runtime,tab:runtime} we compare the runtime and VRAM usage. At an input length of 500 frames, \ours~is up to 93$\times$ faster and has up to 14$\times$ lower VRAM usage than other models. Only VGGT-$\Omega_{256}$ comes close in runtime, at a large cost in accuracy.

\begin{table}[t]
\centering
\caption{\textbf{Model efficiency and quality.} Parameter count, peak inference GPU memory, runtime, and average AUC@30$^\circ$ on the benchmarks of \cref{tab:mvrelpose}, measured on a GH200 with 500 frames. }
\label{tab:runtime}
\small

\begin{tabular}{l rrrr}
\toprule
\textbf{Method}
& \textbf{Params}
& \textbf{Peak Mem}
& \textbf{Time}
& \textbf{AUC30} \\
& (M) $\downarrow$ & (GiB) $\downarrow$ & (s) $\downarrow$ & (\%) $\uparrow$ \\
\midrule

\dejaview & \best 117 & \good 22.4 & 45.0 & 88.8 \\
DA3-G & 1377 & 74.7 & 109.2 & 88.1 \\
$R^3$ & \good 373 & 56.9 & \good 32.0 & 77.9 \\
$\pi^3$ & 959 & 28.8 & 113.3 & \best 91.9 \\
VGGT & 1257 & 48.5 & 130.1 & 82.3 \\
VGGT-$\Omega_{512}$ & 1112 & 41.8 & 51.8 & \good 89.2 \\
VGGT-$\Omega_{256}$ & 1112 & \better 13.9 & \better 4.7 & 78.1 \\
\ours~(ours) & \better 120 & \best 5.3 & \best 1.4 & \better 90.3 \\

\bottomrule
\end{tabular}
\end{table}

\subsection{\ours~Meets Relative Pose Estimation}\label{subsec:results:mvrelpose}

\begin{figure}[t]
\centering
\includegraphics[width=0.99\linewidth]{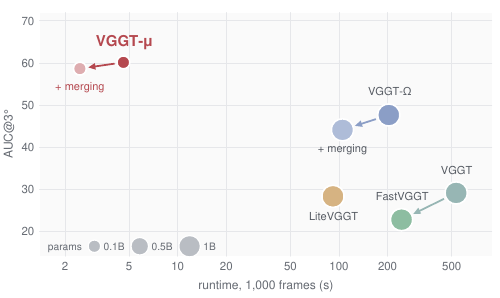}
\caption{\textbf{Accuracy vs.\ runtime on long sequences.} Pose estimation on Cambridge with 1,000 frames. Arrows indicate training-free token merging following \citet{shen2026fastvggt}.}
\label{fig:pareto-long}
\end{figure}

Following~\cite{wang2025pi3}, we evaluate multi-view relative pose estimation. We report Area Under the Curve (AUC) on RealEstate10K (RE10K)~\cite{realestate10k}, ScanNet++~\cite{yeshwanth2023scannet++}, and MegaDepth~\cite{li2018megadepth} and present the results in \cref{tab:mvrelpose}. \ours~achieves accuracy comparable to the baselines but is substantially better than VGGT-$\Omega_{256}$. We additionally compare efficient VGGT variants~\cite{shen2026fastvggt,shu2026litevggtboostingvanillavggt} on long sequences in \cref{fig:pareto-long}. On 1,000 frames inputs, \ours~is $20\times$ faster than the fastest of them and at least 12 AUC@$3^\circ$ points more accurate. The full comparison is in \cref{tab:efficient-vggt} (Suppl.).

\begin{table}[t]
\centering
\caption{\textbf{Multi-view relative pose.} AUC on 10-frame sequences.}
\label{tab:mvrelpose}
\small

\begin{tabular}{l rr rr rr}
\toprule
\textbf{Method} 
& \multicolumn{2}{c}{\textbf{RE10K}}
& \multicolumn{2}{c}{\textbf{ScanNet++}}
& \multicolumn{2}{c}{\textbf{MegaDepth}} \\
\cmidrule(lr){2-3}  
\cmidrule(lr){4-5}  
\cmidrule(lr){6-7}  
AUC@$\rightarrow$ & $3^\circ$ & $30^\circ$ & $3^\circ$ & $30^\circ$ & $3^\circ$ & $30^\circ$ \\
\midrule

\dejaview & 50.0 & 89.0 & \best 69.9 & \best 94.7 & 38.4 & 82.7 \\
DA3-G & \better 57.6 & \good 89.9 & 30.3 & 84.6 & \good 62.9 & 89.9 \\
$R^3$ & 35.9 & 82.3 & 24.9 & 75.0 & 17.9 & 76.4 \\
$\pi^3$ & 54.6 & 88.4 & \best 69.9 & \better 93.5 & \best 68.0 & \best 93.8 \\
VGGT & 25.8 & 79.3 & 18.1 & 74.2 & \better 66.6 & \better 93.5 \\
VGGT-$\Omega_{512}$ & \best 59.2 & \better 90.1 & 26.5 & 84.1 & 62.8 & \good 93.2 \\
VGGT-$\Omega_{256}$ & 45.6 & 87.1 & 2.5 & 62.6 & 32.6 & 84.7 \\
\ours~(ours) & \good 55.4 & \best 92.3 & \good 53.7 & \good 90.2 & 51.9 & 88.2 \\

\bottomrule
\end{tabular}
\end{table}

\subsection{\ours~Meets Visual Localization}\label{subsec:results:visloc}
We evaluate \ours~for visual localization without an explicit 3D map. We first localize each query using multiple retrieved reference images with known poses, then consider localization from a single reference image (see \cref{suppl:single-ref-visloc}, Suppl.). We evaluate calibrated and uncalibrated localization on Cambridge Landmarks~\cite{kendall2015posenet} and Indoor6~\cite{do2022learning}, retrieving $10$ references per query with NetVLAD~\cite{arandjelovic2016netvlad} as in Reloc3r~\cite{reloc3r:2025}. \ours~directly predicts the query pose conditioned on reference poses, additionally using intrinsics for both query and reference images in the calibrated setting. Baseline details are in \cref{suppl:visloc} (Suppl.).

\Cref{tab:visloc_cambridge_indoor6} shows that, when conditioned on reference poses and intrinsics, \ours$_{\mathbf{T}}$ approaches hloc~\cite{sarlin2019coarse} without requiring a pre-built SfM map or local feature matching. It runs about $6\times$ faster with lower peak GPU memory usage. With unknown query focal length, \ours$_\mathbf{T}$ achieves the best rotation errors and competitive translation errors.

\begin{table}[t]
    \centering
\caption{\textbf{Visual localization on Cambridge and Indoor6.} Mean per-scene median translation/rotation errors and accuracy within 0.5\,m, 5$^\circ$ (Cambridge) and 10\,cm, 10$^\circ$ (Indoor6). Subscripts $\mathbf T$/$\mathbf f$ denote conditioning on the reference poses/the focal lengths.
}
    \label{tab:visloc_cambridge_indoor6}
    \small
    \setlength{\tabcolsep}{2.75pt}
    \begin{tabular}{@{}lrrrrrrrr@{}}
        \toprule[1pt]
        \textbf{Method} & \multicolumn{3}{c}{\textbf{Cambridge}} & \multicolumn{3}{c}{\textbf{Indoor6}} & \multicolumn{2}{c}{\textbf{Cost}} \\
        \cmidrule(lr){2-4}\cmidrule(lr){5-7}\cmidrule(lr){8-9}
         & m $\downarrow$ & $^\circ$ $\downarrow$ & \% $\uparrow$ & cm $\downarrow$ & $^\circ$ $\downarrow$ & \% $\uparrow$ & ms $\downarrow$ & GiB $\downarrow$ \\
        \midrule
        \textcolor{gray}{hloc SP+LG} & \textcolor{gray}{0.11} & \textcolor{gray}{0.21} & \textcolor{gray}{90.7} & \textcolor{gray}{1.1} & \textcolor{gray}{0.19} & \textcolor{gray}{94.5} & \textcolor{gray}{250.7} & \textcolor{gray}{0.73} \\
        \midrule
        \multicolumn{9}{l}{\textit{Calibrated — known query focal length}} \\
        $\pi^3_{\mathbf{T}, \mathbf{f}}$ & \better{0.25} & \good{1.49} & \better{76.6} & \better{3.5} & \good{1.16} & \better{87.6} & \better{414.7} & \better{8.86} \\
        MapAny.$_{\mathbf{T}, \mathbf{f}}$ & \good{0.31} & \better{0.75} & \good{76.0} & \good{5.4} & \better{0.96} & \good{81.4} & \good{778.2} & \good{9.61} \\
        \ours$_{\mathbf{T}, \mathbf{f}}$ & \best{0.18} & \best{0.31} & \best{87.1} & \best{2.1} & \best{0.38} & \best{95.0} & \best{45.1} & \best{0.61} \\
        \midrule
        \multicolumn{9}{l}{\textit{Uncalibrated — unknown query focal length}} \\
        Reloc3r$_{512}$ & 0.57 & \better{0.54} & 57.4 & 9.0 & \good{0.78} & 59.7 & 145.8 & \good{2.70} \\
        \dejaview & \best{0.28} & 2.08 & \better{69.5} & \best{3.3} & 1.75 & \good{81.0} & 144.2 & \better{2.30} \\
        DA3-G & 0.37 & 2.68 & 57.3 & \better{3.5} & 1.71 & 77.1 & 192.2 & 8.14 \\
        $R^3$ & 0.41 & \good{0.55} & 64.0 & 4.9 & \better{0.70} & 74.7 & \good{101.0} & 3.03 \\
        $\pi^3_{\mathbf{T}}$ & 0.36 & 1.93 & 65.6 & 4.9 & 1.63 & 80.3 & 408.6 & 8.86 \\
        VGGT-$\Omega_{512}$ & \good{0.32} & 2.29 & \good{66.0} & \better{3.5} & 1.67 & \better{81.3} & 182.2 & 6.00 \\
        VGGT-$\Omega_{256}$ & 0.37 & 2.47 & 62.1 & 4.8 & 1.93 & 76.7 & \better{71.5} & 4.96 \\
        MapAny.$_{\mathbf{T}}$ & 0.44 & 0.88 & 59.6 & 7.7 & 1.29 & 66.0 & 770.7 & 9.59 \\
        \ours$_{\mathbf{T}}$ & \better{0.31} & \best{0.32} & \best{74.0} & \good{4.1} & \best{0.43} & \best{84.3} & \best{46.8} & \best{0.61} \\
        \bottomrule[1pt]
    \end{tabular}%
\end{table}

\subsection{\ours~Meets SfM}\label{subsec:results:sfm}
\ours~can replace the global stage of a classical SfM pipeline. A single forward pass followed by anisotropic pose averaging gives absolute poses and focal lengths for all images of a scene, which replace rotation averaging and global positioning. Under these poses, we triangulate the same verified matches that GLOMAP uses and refine the reconstruction with alternating bundle adjustment, track merging and outlier filtering (details in \cref{suppl:sfm}, Suppl.). \Cref{tab:global_sfm} reports the mean over five IMC23 Phototourism scenes~\cite{image-matching-challenge-2023}. Our method is slightly more accurate than GLOMAP at all thresholds, and is far ahead of incremental COLMAP. Including \ours~inference, it is about $4\times$ faster than GLOMAP and $28\times$ faster than COLMAP.

\begin{table}[t]
  \centering
  \caption{\textbf{Global SfM on IMC23 Phototourism.}
    Mean relative pose AUC (\%) over five scenes (British Museum,
    Pantheon Exterior, Brandenburg Gate, Taj Mahal, and Colosseum
    Exterior), with mean runtime on an RTX 4090.
    Methods are evaluated with exhaustive SIFT~\cite{lowe2004distinctive} matches. Feature extraction and matching are shared
    for all methods and excluded from the timings.
  }
  \label{tab:global_sfm}
  \small
  \setlength{\tabcolsep}{4pt}
  \begin{tabular}{@{}lccccr@{}}
    \toprule
    \textbf{Method} & \multicolumn{4}{c}{\textbf{AUC} (\%) $\uparrow$}
    & \\
    \cmidrule(lr){2-5}
    AUC@$\rightarrow$ & @0.5\textdegree
    & @1\textdegree
    & @5\textdegree
    & @10\textdegree
    & Time (s) $\downarrow$ \\
    \midrule
    COLMAP~\cite{schoenberger2016sfm}
      & 12.3 & 25.7 & 61.9 & 73.6 & 2290 \\
    GLOMAP~\cite{pan2024glomap}
      & \better{40.5} & \better{55.6}
      & \better{80.5} & \better{86.2} & \better{358} \\
    \ours~(ours)
      & \best{40.6} & \best{56.2}
      & \best{81.3} & \best{87.0} & \best{81} \\
    \bottomrule
  \end{tabular}
\end{table}

\subsection{\ours~Meets Visual Odometry}\label{subsec:results:vo}
We use \ours~for VO by processing the stream in overlapping keyframe blocks, localizing every non-keyframe in a single pass against the most recent block, reusing cached keys and values, and connecting blocks through a scale-aware pose graph. Periodic joint passes over distant keyframes provide loop closures. Implementation details and evaluation settings are provided in \cref{suppl:vo} (Suppl.).

We benchmark on Cambridge Landmarks~\cite{kendall2015posenet} and ETH3D SLAM~\cite{schops2019bad}. \Cref{tab:odometry} shows that \ours~is competitive in terms of FPS, latency, and memory usage on both datasets,
and second-most accurate in terms of absolute trajectory error (ATE). All other methods have drawbacks: DROID-SLAM~\cite{teed2021droid} is slow, has high VRAM usage, and can produce outlier errors that cause high ATEs. ORB-SLAM3~\cite{ORBSLAM3_TRO} fails on multiple scenes. VGGT-SLAM 2.0~\cite{maggio2026vggt2} has high latency and VRAM usage.
In \cref{suppl:vo-low-res} (Suppl.), we provide additional metrics and run all methods at the same low resolution as \ours. There, ORB-SLAM3 and VGGT-SLAM~2.0 fail frequently, and DROID-SLAM is much slower than the other methods.

\begin{table}[t]
\centering
\caption{\textbf{Visual odometry on Cambridge and ETH3D SLAM.} Success (evaluable sequences per run), FPS (RTX 4090), latency post-initialization  (ms), peak VRAM (GiB), and ATE (m).
}
\label{tab:odometry}

\small
\setlength{\tabcolsep}{4pt}
\begin{tabular}{@{}l rrrrr@{}}
\toprule
\textbf{Method}
& Succ.$\uparrow$ & FPS$\uparrow$ & Lat.$\downarrow$
& VRAM$\downarrow$ & ATE$\downarrow$ \\
\midrule

\multicolumn{6}{@{}l}{\textit{Cambridge Landmarks}} \\
DROID-SLAM
& \best 8/8 & 14.9 & 61.3 & \better 5.0 & 7.00 \\
ORB-SLAM3
& 5/8 & \better 42.1 & \better 27.8 & n/a & \best 0.78 \\
VGGT-SLAM 2.0
& \best 8/8 & 26.6 & 840.9 & 5.5 & 1.36 \\
\ours~(ours)
& \best 8/8 & \best 67.1 & \best 16.4 & \best 0.6 & \better 1.13 \\
\midrule

\multicolumn{6}{@{}l}{\textit{ETH3D SLAM}} \\
DROID-SLAM
& \best 6/6 & 37.2 & 54.4 & \better 3.8 & \best 0.39 \\
ORB-SLAM3
& 5/6 & \best 221.5 & \best 5.3 & n/a & 0.73 \\
VGGT-SLAM 2.0
& \best 6/6 & \better 213.1 & 800.0 & 5.4 & 0.68 \\
\ours~(ours)
& \best 6/6 & 94.5 & \better 10.6 & \best 0.7 & \better 0.54 \\
\bottomrule
\end{tabular}
\end{table}

\subsection{\ours~Meets Scene Clustering}\label{subsec:results:mvrelpose-mixing}

We evaluate pose estimation with inputs mixing three scenes of eight frames each. \Cref{tab:mixed-scene} reports pose accuracy and \cref{fig:teaser}(c) shows the effect of scene mixing and distractors (see \cref{suppl:scene-clustering,tab:mixed-drop}, Suppl.). \ours~needs no clustering and suffers only a small drop under scene mixing. \Cref{tab:scene-separation} evaluates clustering with a single threshold per method, tuned on separate mixes. \ours~($\gamma=0.5$) achieves better separation than retrieval-based methods.

\begin{table}[t]
\centering
\caption{\textbf{Pose estimation on disconnected view graphs.}
Each input contains 8 frames from each of 3 unrelated scenes, shuffled. We compare running all in one forward pass to first running MegaLoc~\cite{berton2025megaloc} and reconstructing each cluster separately. \ours~clusters with $\hat{c}_{ij}$. More baselines are in \cref{tab:mixed-scene-full} (Suppl.).}
\label{tab:mixed-scene}

\small
\setlength{\tabcolsep}{3pt}
\begin{tabular}{@{}l c rr rr rr@{}}
\toprule
\textbf{Method}
& \textbf{MegaLoc}
& \multicolumn{2}{c}{\textbf{RE10K}}
& \multicolumn{2}{c}{\textbf{ScanNet++}}
& \multicolumn{2}{c}{\textbf{MegaDepth}} \\
\cmidrule(lr){3-4}
\cmidrule(lr){5-6}
\cmidrule(lr){7-8}
AUC@$\rightarrow$ &
& $3^\circ$ & $30^\circ$
& $3^\circ$ & $30^\circ$
& $3^\circ$ & $30^\circ$ \\
\midrule

\multicolumn{8}{@{}l}{\textit{Single forward pass}} \\
$\pi^3$              & \xmark & \better{15.2} & \better{57.1} & \better{19.5} & \better{51.8} & \better{40.0} & 82.1 \\
VGGT-$\Omega$  & \xmark & 13.9 & 56.2 &  5.7 & 47.8 & 33.9 & \better{83.0} \\
\ours                & \xmark & \best{52.1} & \best{91.2} & \best{42.0} & \best{83.5} & \best{49.0} & \best{86.9} \\
\midrule
\multicolumn{8}{@{}l}{\textit{Cluster, then reconstruct}} \\
$\pi^3$              & \cmark & 52.8 & 87.1 & \best{57.6} & \better{81.3} & \best{67.2} & \best{92.6} \\
VGGT-$\Omega$  & \cmark & \best{57.6} & \better{88.9} & 18.7 & 71.4 & \better{61.0} & \better{91.7} \\
\ours                & \xmark & \better{54.7} & \best{92.0} & \better{47.8} & \best{86.4} & 51.7 & 87.7 \\
\bottomrule
\end{tabular}
\end{table}

\begin{table}[t]
\centering
\caption{\textbf{Scene separation.}
Same inputs as \cref{tab:mixed-scene}. We compare clustering using retrieval methods' global descriptors and \ours~using predicted edge confidence $\hat{c}_{ij}$. We report Adjusted Rand Index (ARI) and Success Rate (SR), the fraction of mixes with all scenes recovered exactly. More baselines are in \cref{tab:scene-separation-full} (Suppl.).}
\label{tab:scene-separation}
\small
\setlength{\tabcolsep}{4pt}
\begin{tabular}{@{}l rr rr rr@{}}
\toprule
\textbf{Method}
& \multicolumn{2}{c}{\textbf{RE10K}}
& \multicolumn{2}{c}{\textbf{ScanNet++}}
& \multicolumn{2}{c}{\textbf{MegaDepth}} \\
\cmidrule(lr){2-3}\cmidrule(lr){4-5}\cmidrule(lr){6-7}
& ARI & SR & ARI & SR & ARI & SR \\
\midrule
NetVLAD~\cite{arandjelovic2016netvlad}   & \better{93.2} & {77.5} & {67.4} & {2.0} & 66.6 & 0.0 \\
DINOv2~\cite{oquab2023dinov2}            & 88.3 & 65.0 & 53.5 & 0.5 & {96.1} & {75.5} \\
MegaLoc~\cite{berton2025megaloc}         & {91.8} & \better{81.5} & \better{86.7} & \better{31.0} & \better{98.1} & \better{80.0} \\
\ours                                    & \best{99.5} & \best{98.5} & \best{96.4} & \best{83.0} & \best{98.6} & \best{94.0} \\
\bottomrule
\end{tabular}
\end{table}

\subsection{Ablations}\label{subsec:results:ablations}

\paragraph{Anisotropic averaging.}
\Cref{tab:anisotropic-averaging} compares methods for recovering absolute poses from the same edges by greedy chaining~\cite{xu2026r33dreconstructionrelative}, which places each frame by a confidence-weighted average of its poses relative to already placed frames, and by our pose averaging (\cref{sec:pose-avr}) with equal edge weights (isotropic) or weights $\hat{\boldsymbol{\Lambda}}_{ij}$ (anisotropic). Averaging improves over chaining, and $\hat{\boldsymbol{\Lambda}}_{ij}$ improves it further, most notably at strict thresholds.

\begin{table}[t]
    \centering
    \caption{\textbf{Pose averaging for visual localization on Indoor6.}
    Each solver converts the same predicted relative poses into
    absolute query poses using known reference poses. We report mean per-scene median errors and accuracy.}
    \label{tab:anisotropic-averaging}
    \small
    \renewcommand{\arraystretch}{1.12}
    \setlength{\tabcolsep}{4pt}
    \begin{tabular}{l rr @{\hspace{7pt}} rr}
        \toprule
        \textbf{Averaging} 
        & \multicolumn{2}{c}{\textbf{Pose error} $\downarrow$}
        & \multicolumn{2}{c}{\textbf{Accuracy} (\%) $\uparrow$} \\
        \cmidrule(lr){2-3}
        \cmidrule(lr){4-5}
        & cm & $^\circ$
        & 5\,cm, $5^\circ$
        & 10\,cm, $10^\circ$ \\
        \midrule
        Chaining
        & 4.30 & 0.456
        & 60.48 & 83.65 \\
        Isotropic
        & \better{4.21} & \better{0.438}
        & \better{61.12} & \better{83.73} \\
        Anisotropic
        & \best{4.09} & \best{0.430}
        & \best{61.95} & \best{84.33} \\
        \bottomrule
    \end{tabular}
\end{table}

\paragraph{Model design.} \Cref{tab:arch-ablation} compares model design choices, starting from training only on GT and from last-layer distillation of VGGT-$\Omega$. Per-layer distillation combined with GT supervision substantially improves pose accuracy. Our relative pose formulation further improves AUC@$3^\circ$ on MegaDepth while enabling uncertainty estimation. Looped weight sharing reduces the number of parameters while improving accuracy on RE10K, and restricting training to datasets with depth maps weakens performance. The supplementary further shows that training on mixed scenes improves robustness (\cref{fig:mixed-ablation}) and training at 256 versus 512 pixels gives higher accuracy for the same compute (\cref{fig:iso-compute}).

\begin{table}[t]
\centering
\caption{\textbf{Model design ablations}. Pose accuracy for different model design choices after one day of training on 4 GPUs.}
\label{tab:arch-ablation}
\small
\setlength{\tabcolsep}{3.5pt}
\begin{tabular}{@{}l r c rr rr@{}}
\toprule
\textbf{Method} & \textbf{Params}$^\dagger$ & \textbf{$\hat{c}_{ij}$}
& \multicolumn{2}{c}{\textbf{MegaDepth}}
& \multicolumn{2}{c}{\textbf{RE10K}} \\
\cmidrule(lr){4-5}\cmidrule(lr){6-7}
AUC@$\rightarrow$ & & & $3^\circ$ & $30^\circ$ & $3^\circ$ & $30^\circ$ \\
\midrule
Train only on GT & 157M & \xmark & 9.8 & 68.0 & 0.8 & 22.4 \\
VGGT-$\Omega$ distill      & 157M & \xmark & 3.5 & 56.1 & 2.3 & 40.9 \\
\quad + distill per layer         & 157M & \xmark & 15.3 & 76.5 & 11.0 & 67.1 \\
\quad + GT supervision                  & 157M & \xmark & 27.1 & \best{83.3} & \better{14.8} & \better{71.8} \\
\quad + relative pose                   & 159M & \cmark & \best{34.9} & \better{82.8} & \good{14.5} & \good{68.0} \\
\rowcolor{gray!25} \quad + weight sharing           & \textbf{120M} & \cmark & \better{34.2} & \good{82.6} & \best{19.2} & \best{73.1} \\
\quad - pose-only data           & \textbf{120M} & \cmark & \good{32.7} & 81.7 & 11.2 & 65.9 \\
\bottomrule
\end{tabular}
\end{table}

\section{Conclusion, Limitations and Future Work}

We introduced \ours, a lightweight relative pose estimator designed for efficient inference at scale. With only 120M parameters, it rivals its much larger counterpart, VGGT-$\Omega$, on pose benchmarks while achieving strong downstream performance in visual localization, SfM, and VO. \ours~further enables robust estimation on unordered image collections containing unrelated scenes and predicts pose information matrices for uncertainty-aware pose averaging. Finally, it supports metric pose estimation and conditioning on known camera parameters, making it a flexible building block for downstream geometric vision tasks.

One limitation is that the predicted $6\times6$ information matrices provide limited accuracy gains. Still, their approximate calibration makes them promising for downstream sensor fusion~\cite{pataki2026vidmap}, which we leave to future work. While anisotropic pose averaging assumes independent edge uncertainties, our multi-view features make them correlated. Predicting uncertainties that are independent constitute interesting future work.

Another limitation is that \ours~does not predict dense geometry, which limits its use to pose related tasks. In addition, the low-resolution ($256\times256$) input costs some accuracy at strict thresholds such as AUC@$3^\circ$ (\cref{tab:mvrelpose}). %

\section*{LLM Usage Statement}
We used Claude AI (Anthropic), Claude Code (Anthropic), ChatGPT (OpenAI), and Codex (OpenAI) to assist with:
\begin{itemize}
    \item Generating code for the experiments in \cref{sec:results}.
    \begin{itemize}
        \item Source code was read by the authors and evaluations were based on previous work.
        \item The core implementation of the model and loss was done with careful auditing by the authors.
    \end{itemize}
    \item Checking for typos.
    \item Identifying relevant literature (checked by the authors).
    \item Generating the TikZ and Matplotlib code for the figures.
    \item Assisting in designing experiments.
\end{itemize}

\paragraph{We have reviewed all LLM-assisted work.} We take responsibility for the final content of this work, including text, claims or artifacts produced with the aid of generative AI.

\section*{Acknowledgements}
This work was supported by the Wallenberg Artificial
Intelligence, Autonomous Systems and Software Program
(WASP), funded by the Knut and Alice Wallenberg Foundation, the Crafoord Foundation, and by the strategic research environment ELLIIT, funded by the Swedish government. 
The computational resources were provided by the
National Academic Infrastructure for Supercomputing in
Sweden (NAISS) at C3SE, partially funded by the Swedish Research
Council through grant agreement no.~2022-06725, and by
the Berzelius resource, provided by the Knut and Alice Wallenberg Foundation at the National Supercomputer Centre.

{
    \small
    \bibliographystyle{ieeenat_fullname}
    \bibliography{main}

@String(PAMI = {IEEE Trans. Pattern Anal. Mach. Intell.})

@String(IJCV = {Int. J. Comput. Vis.})

@String(CVPR= {IEEE Conf. Comput. Vis. Pattern Recog.})

@String(ICCV= {Int. Conf. Comput. Vis.})

@String(ECCV= {Eur. Conf. Comput. Vis.})

@String(NIPS= {Adv. Neural Inform. Process. Syst.})

@String(BMVC= {Brit. Mach. Vis. Conf.})

@String(TOG= {ACM Trans. Graph.})

@String(ICLR = {Int. Conf. Learn. Represent.})

@String(CVPRW= {IEEE Conf. Comput. Vis. Pattern Recog. Worksh.})

@String(PAMI  = {IEEE TPAMI})

@String(IJCV  = {IJCV})

@String(CVPR  = {CVPR})

@String(ICCV  = {ICCV})

@String(ECCV  = {ECCV})

@String(NIPS  = {NeurIPS})

@String(BMVC  =	{BMVC})

@String(TOG   = {ACM TOG})

@String(ICLR  = {ICLR})

@String(CVPRW= {CVPRW})

@inproceedings{adamw:2019,
  title={Decoupled Weight Decay Regularization},
  author={Ilya Loshchilov and Frank Hutter},
  booktitle=iclr,
  year={2019}}

@INPROCEEDINGS{SuperPoint:2018,
  author={DeTone, Daniel and Malisiewicz, Tomasz and Rabinovich, Andrew},
  booktitle=cvprw, 
  title={SuperPoint: Self-Supervised Interest Point Detection and Description}, 
  year={2018}
}

@book{Hartley_Zisserman_2004, place={Cambridge}, edition={2nd}, title={Multiple View Geometry in Computer Vision}, publisher={Cambridge University Press}, author={Hartley, Richard and Zisserman, Andrew}, year={2004}}

@InProceedings{reloc3r:2025,
    author    = {Dong, Siyan and Wang, Shuzhe and Liu, Shaohui and Cai, Lulu and Fan, Qingnan and Kannala, Juho and Yang, Yanchao},
    title     = {Reloc3r: Large-Scale Training of Relative Camera Pose Regression for Generalizable, Fast, and Accurate Visual Localization},
    booktitle = cvpr,
    year      = {2025},
    pages     = {16739-16752}
}

@inproceedings{sarlin2020superglue,
  title={Superglue: Learning feature matching with graph neural networks},
  author={Sarlin, Paul-Edouard and DeTone, Daniel and Malisiewicz, Tomasz and Rabinovich, Andrew},
  booktitle=cvpr,
  year={2020}
}

@article{realestate10k,title	= {Stereo Magnification: Learning view synthesis using multiplane images},author	= {Tinghui Zhou and Richard Tucker and John Flynn and Graham Fyffe and Noah Snavely},year	= {2018},URL	= {https://arxiv.org/abs/1805.09817},journal	= tog,volume	= {37}}

@inproceedings{wang2025pi3,
      title={$\pi^3$: Permutation-Equivariant Visual Geometry Learning}, 
      author={Yifan Wang and Jianjun Zhou and Haoyi Zhu and Wenzheng Chang and Yang Zhou and Zizun Li and Junyi Chen and Jiangmiao Pang and Chunhua Shen and Tong He},
      year={2026},
      booktitle=iclr
}

@article{cabon2020vkitti2,
      title={Virtual {KITTI} 2}, 
      author={Yohann Cabon and Naila Murray and Martin Humenberger},
      journal={arXiv preprint arXiv:2001.10773},
      year={2020},
      eprint={2001.10773},
      archivePrefix={arXiv},
      primaryClass={cs.CV},
      url={https://arxiv.org/abs/2001.10773}, 
}

@inproceedings{keetha2026mapanything,
  title={{MapAnything}: Universal Feed-Forward Metric {3D} Reconstruction},
  author={Nikhil Keetha and Norman M\"{u}ller and Johannes L. Sch\"{o}nberger and Lorenzo Porzi and Yuchen Zhang and Tobias Fischer and Arno Knapitsch and Duncan Zauss and Ethan Weber and Nelson Antunes and Jonathon Luiten and Manuel L{\'o}pez-Antequera and Samuel {Rota Bul\`{o}} and Christian Richardt and Deva Ramanan and Sebastian Scherer and Peter Kontschieder},
  booktitle={3DV},
  year={2026}
}

@inproceedings{arandjelovic2016netvlad,
  title={NetVLAD: CNN Architecture for Weakly Supervised Place Recognition},
  author={Arandjelovi{\'c}, Relja and Gronat, Petr and Torii, Akihiko and Pajdla, Tomas and Sivic, Josef},
  booktitle=cvpr,
  year={2016}
}

@inproceedings{li2026longtail,
  title={Long-Tail Internet Photo Reconstruction},
  author={Li, Yuan and Xiangli, Yuanbo and Averbuch-Elor, Hadar and Snavely, Noah and Cai, Ruojin},
  booktitle=cvpr,
  year={2026}
}

@inproceedings{durso202Xterrasky3d,
  title={TerraSky3D: Multi-View Reconstructions of European Landmarks in 4K},
    author={D'Urso, Mattia and Hu, Yuxi and Rossi, Mattia and Sormann, Christian and Fraundorfer, Friedrich},
  booktitle=cvprw,
  year={2026}
}

@inproceedings{schoenberger2016sfm,
    title={{Structure-from-Motion Revisited}},
    author={Sch{\"o}nberger, Johannes L. and Frahm, Jan-Michael},
    booktitle=cvpr,
    year={2016},
}

@inproceedings{wang2024dust3r,
  title={DUSt3R: Geometric 3D Vision Made Easy},
  author={Wang, Shuzhe and Leroy, Vincent and Cabon, Yohann and Chidlovskii, Boris and Revaud, J{\'e}r{\^o}me},
  booktitle=cvpr,
  year={2024}
}

@inproceedings{wang2025vggt,
  title={VGGT: Visual Geometry Grounded Transformer},
  author={Wang, Jianyuan and Chen, Minghao and Karaev, Nikita and Vedaldi, Andrea and Rupprecht, Christian and Novotny, David},
  booktitle=cvpr,
  year={2025}
}

@article{Zhao2023ALIKED,
    title = {ALIKED: A Lighter Keypoint and Descriptor Extraction Network via Deformable Transformation},
    author = {Zhao, Xiaoming and Wu, Xingming and Chen, Weihai and Chen, Peter C. Y. and Xu, Qingsong and Li, Zhengguo},
    journal = {IEEE Transactions on Instrumentation \& Measurement},
    volume = {72},
    year = {2023},
}

@InProceedings{Izquierdo_CVPR_2024_SALAD,
    author    = {Izquierdo, Sergio and Civera, Javier},
    title     = {Optimal Transport Aggregation for Visual Place Recognition},
    booktitle = cvpr,
    month     = {June},
    year      = {2024},
}

@inproceedings{nordstrom2026lomalocalfeaturematching,
      title={LoMa: Local Feature Matching Revisited}, 
      author={David Nordstr\"om and Johan Edstedt and Georg B\"okman and Jonathan Astermark and Anders Heyden and Viktor Larsson and M{\aa}rten Wadenb{\"a}ck and Michael Felsberg and Fredrik Kahl},
      booktitle=eccv,
      year={2026}
}

@article{lowe2004distinctive,
  title={{Distinctive image features from scale-invariant keypoints}},
  author={Lowe, David G.},
  journal=ijcv,
  year={2004}
}

@inproceedings{sarlin2019coarse,
  title={{From coarse to fine: Robust hierarchical localization at large scale}},
  author={Sarlin, Paul-Edouard and Cadena, Cesar and Siegwart, Roland and Dymczyk, Marcin},
  booktitle=cvpr,
  year={2019}
}

@article{umeyama1991leastsquares,
  author={Umeyama, Shinji},
  title={Least-squares estimation of transformation parameters between two point patterns}, 
  journal=PAMI, 
  year={1991},
  volume={13},
  number={4},
}

@inproceedings{li2018megadepth,
  title={Megadepth: Learning single-view depth prediction from internet photos},
  author={Li, Zhengqi and Snavely, Noah},
  booktitle=cvpr,
  year={2018}
}

@inproceedings{kendall2015posenet,
  title={Posenet: A convolutional network for real-time 6-dof camera relocalization},
  author={Kendall, Alex and Grimes, Matthew and Cipolla, Roberto},
  booktitle=iccv,
  year={2015}
}

@article{mur2015orb,
  title={{ORB-SLAM}: a versatile and accurate monocular SLAM system},
  author={Mur-Artal, Raul and Montiel, Jose Maria Martinez and Tardos, Juan D},
  journal={IEEE Transactions on Robotics},
  volume={31},
  number={5},
  pages={1147--1163},
  year={2015},
  publisher={IEEE}
}

@inproceedings{edstedt2023dkm,
  title={{DKM}: Dense Kernelized Feature Matching for Geometry Estimation},
  author={Edstedt, Johan and Athanasiadis, Ioannis and Wadenb{\"a}ck, M{\aa}rten and Felsberg, Michael},
  booktitle=cvpr,
  year={2023}
}

@article{oquab2023dinov2,
  title={{DINOv2}: Learning Robust Visual Features without Supervision},
  author={Oquab, Maxime and Darcet, Timoth{\'e}e and Moutakanni, Theo and Vo, Huy V. and Szafraniec, Marc and Khalidov, Vasil and Fernandez, Pierre and Haziza, Daniel and Massa, Francisco and El-Nouby, Alaaeldin and Howes, Russell and Huang, Po-Yao and Xu, Hu and Sharma, Vasu and Li, Shang-Wen and Galuba, Wojciech and Rabbat, Mike and Assran, Mido and Ballas, Nicolas and Synnaeve, Gabriel and Misra, Ishan and Jegou, Herve and Mairal, Julien and Labatut, Patrick and Joulin, Armand and Bojanowski, Piotr},
  journal={Transactions on Machine Learning Research},
  year={2024}
}

@inproceedings{edstedt2024roma,
  title={{RoMa}: Robust Dense Feature Matching},
  author={Edstedt, Johan and Sun, Qiyu and B{\"o}kman, Georg and Wadenb{\"a}ck, M{\aa}rten and Felsberg, Michael},
  booktitle=cvpr,
  year={2024},
}

@article{zhao2022alike,
  title={Alike: Accurate and lightweight keypoint detection and descriptor extraction},
  author={Zhao, Xiaoming and Wu, Xingming and Miao, Jinyu and Chen, Weihai and Chen, Peter CY and Li, Zhengguo},
  journal={IEEE Transactions on Multimedia},
  year={2022},
}

@inproceedings{brynte2024learning,
  title={Learning structure-from-motion with graph attention networks},
  author={Brynte, Lucas and Iglesias, Jos{\'e} Pedro and Olsson, Carl and Kahl, Fredrik},
  booktitle=cvpr,
  year={2024},
}

@inproceedings{pan2024glomap,
    author={Pan, Linfei and Barath, Daniel and Pollefeys, Marc and Sch\"{o}nberger, Johannes L.},
    title={{Global Structure-from-Motion Revisited}},
    booktitle=eccv,
    year={2024},
}

@inproceedings{leroy2024grounding,
  title={Grounding Image Matching in 3D with MASt3R},
  author={Leroy, Vincent and Cabon, Yohann and Revaud, J{\'e}r{\^o}me},
  booktitle=eccv,
  year={2024},
}

@inproceedings{jang2025pow3r,
  title={Pow3r: Empowering Unconstrained 3D Reconstruction with Camera and Scene Priors},
  author={Jang, Wonbong and Weinzaepfel, Philippe and Leroy, Vincent and Agapito, Lourdes and Revaud, Jerome},
  booktitle=cvpr,
  year={2025}
}

@inproceedings{tung2024megascenes,
  title={Megascenes: Scene-Level View Synthesis at Scale},
  author={Tung, Joseph and Chou, Gene and Cai, Ruojin and Yang, Guandao and Zhang, Kai and Wetzstein, Gordon and Hariharan, Bharath and Snavely, Noah},
  booktitle=eccv,
  year={2024},
}

@inproceedings{berton2023eigenplaces,
  title={Eigenplaces: Training viewpoint robust models for visual place recognition},
  author={Berton, Gabriele and Trivigno, Gabriele and Caputo, Barbara and Masone, Carlo},
  booktitle=iccv,
  year={2023}
}

@inproceedings{berton2025megaloc,
  title={Megaloc: One retrieval to place them all},
  author={Berton, Gabriele and Masone, Carlo},
  booktitle=cvprw,
  year={2025}
}

@inproceedings{vuong2025aerialmegadepth,
  title={Aerialmegadepth: Learning aerial-ground reconstruction and view synthesis},
  author={Vuong, Khiem and Ghosh, Anurag and Ramanan, Deva and Narasimhan, Srinivasa and Tulsiani, Shubham},
  booktitle=cvpr,
  year={2025}
}

@inproceedings{roberts2021hypersim,
    author    = {Mike Roberts AND Jason Ramapuram AND Anurag Ranjan AND Atulit Kumar AND
                 Miguel Angel Bautista AND Nathan Paczan AND Russ Webb AND Joshua M. Susskind},
    title     = {{Hypersim}: {A} Photorealistic Synthetic Dataset for Holistic Indoor Scene Understanding},
    booktitle = iccv,
    year      = {2021}
}

@inproceedings{ling2024dl3dv,
  title={Dl3dv-10k: A large-scale scene dataset for deep learning-based 3d vision},
  author={Ling, Lu and Sheng, Yichen and Tu, Zhi and Zhao, Wentian and Xin, Cheng and Wan, Kun and Yu, Lantao and Guo, Qianyu and Yu, Zixun and Lu, Yawen and others},
  booktitle=cvpr,
  year={2024}
}

@article{simeoni2025dinov3,
  title={DINOv3}, 
  author={Oriane Sim{\'e}oni and Huy V. Vo and Maximilian Seitzer and Federico Baldassarre and Maxime Oquab and Cijo Jose and Vasil Khalidov and Marc Szafraniec and Seungeun Yi and Micha{\"e}l Ramamonjisoa and Francisco Massa and Daniel Haziza and Luca Wehrstedt and Jianyuan Wang and Timoth{\'e}e Darcet and Th{\'e}o Moutakanni and Leonel Sentana and Claire Roberts and Andrea Vedaldi and Jamie Tolan and John Brandt and Camille Couprie and Julien Mairal and Herv{\'e} J{\'e}gou and Patrick Labatut and Piotr Bojanowski},
  journal={Transactions on Machine Learning Research},
  
  year={2026}
}

@article{panek2026guide,
  title={A Guide to Structureless Visual Localization},
  author={Panek, Vojtech and Zhou, Qunjie and Ding, Yaqing and Agostinho, S{\'e}rgio and Kukelova, Zuzana and Sattler, Torsten and Leal-Taix{\'e}, Laura},
  journal={International Journal of Computer Vision},
  volume={134},
  number={6},
  pages={263},
  year={2026},
  publisher={Springer}
}

@inproceedings{dillen2026sparse,
  title={Sparse-View Localization via Online Neural 3D Regression},
  author={Dill{\'e}n, Ludvig and Oskarsson, Magnus and Larsson, Viktor},
  booktitle=cvpr,
  pages={21794--21804},
  year={2026}
}

@inproceedings{yao2020blendedmvs,
  title={BlendedMVS: A Large-scale Dataset for Generalized Multi-view Stereo Networks},
  author={Yao, Yao and Luo, Zixin and Li, Shiwei and Zhang, Jingyang and Ren, Yufan and Zhou, Lei and Fang, Tian and Quan, Long},
  booktitle=cvpr,
  year={2020}
}

@inproceedings{arnold2022map,
  title={Map-free visual relocalization: Metric pose relative to a single image},
  author={Arnold, Eduardo and Wynn, Jamie and Vicente, Sara and Garcia-Hernando, Guillermo and Monszpart, Aron and Prisacariu, Victor Adrian and Turmukhambetov, Daniyar and Brachmann, Eric},
  booktitle=eccv,
  year={2022}
}

@inproceedings{zhang2025ufm,
 title={UFM: A Simple Path towards Unified Dense Correspondence with Flow},
 author={Zhang, Yuchen and Keetha, Nikhil and Lyu, Chenwei and Jhamb, Bhuvan and Chen, Yutian and Qiu, Yuheng and Karhade, Jay and Jha, Shreyas and Hu, Yaoyu and Ramanan, Deva and Scherer, Sebastian and Wang, Wenshan},
 booktitle=NIPS,
 year={2025}
}

@inproceedings{gaidon2016virtual,
  title={Virtual worlds as proxy for multi-object tracking analysis},
  author={Gaidon, Adrien and Wang, Qiao and Cabon, Yohann and Vig, Eleonora},
  booktitle=cvpr,
  year={2016}
}

@inproceedings{yeshwanth2023scannet++,
  title={Scannet++: A high-fidelity dataset of 3d indoor scenes},
  author={Yeshwanth, Chandan and Liu, Yueh-Cheng and Nie{\ss}ner, Matthias and Dai, Angela},
  booktitle=iccv,
  year={2023}
}

@inproceedings{wang2025spatialvid,
  title={Spatialvid: A large-scale video dataset with spatial annotations},
  author={Wang, Jiahao and Yuan, Yufeng and Zheng, Rujie and Lin, Youtian and Gao, Jian and Chen, Lin-Zhuo and Bao, Yajie and Zeng, Chang and Zhou, Yanxi and Long, Xiao-Xiao and Zhu, Hao and Zhang, Zhaoxiang and Cao, Xun and Yao, Yao},
  booktitle=cvpr,
  year={2026}
}

@InProceedings{AriaSynthEnv:2025,
author="Avetisyan, Armen
and Xie, Christopher
and Howard-Jenkins, Henry
and Yang, Tsun-Yi
and Aroudj, Samir
and Patra, Suvam
and Zhang, Fuyang
and Frost, Duncan
and Holland, Luke
and Orme, Campbell
and Engel, Jakob
and Miller, Edward
and Newcombe, Richard
and Balntas, Vasileios",
editor="Leonardis, Ale{\v{s}}
and Ricci, Elisa
and Roth, Stefan
and Russakovsky, Olga
and Sattler, Torsten
and Varol, G{\"u}l",
title="SceneScript: Reconstructing Scenes with an Autoregressive Structured Language Model",
booktitle=eccv,
year="2024",
}

@inproceedings{mpsd:2020,
  title={Mapillary planet-scale depth dataset},
  author={L{\'o}pez-Antequera, Manuel and Gargallo, Pau and Hofinger, Markus and {Rota Bul\`{o}}, Samuel and Kuang, Yubin and Kontschieder, Peter},
  booktitle=eccv,
  year={2020},
}

@inproceedings{edstedt2026romav2,
  title = {{RoMa} v2: Harder Better Faster Denser Feature Matching},
  author = {Edstedt, Johan and Nordstr{\"o}m, David and Zhang, Yushan and B{\"o}kman, Georg and Astermark, Jonathan and Larsson, Viktor and Heyden, Anders and Kahl, Fredrik and Wadenb{\"a}ck, M{\aa}rten and Felsberg, Michael},
  booktitle = eccv,
  year = {2026}
}

@InProceedings{Jiang_2025_CVPR,
    author    = {Jiang, Hanwen and Xu, Zexiang and Xie, Desai and Chen, Ziwen and Jin, Haian and Luan, Fujun and Shu, Zhixin and Zhang, Kai and Bi, Sai and Sun, Xin and Gu, Jiuxiang and Huang, Qixing and Pavlakos, Georgios and Tan, Hao},
    title     = {MegaSynth: Scaling Up 3D Scene Reconstruction with Synthesized Data},
    booktitle = cvpr,
    year      = {2025},
}

@inproceedings{depthanything3,
  title={Depth Anything 3: Recovering the visual space from any views},
  author={Haotong Lin and Sili Chen and Jun Hao Liew and Donny Y. Chen and Zhenyu Li and Yang Zhao and Sida Peng and Hengkai Guo and Xiaowei Zhou and Guang Shi and Jiashi Feng and Bingyi Kang},
  year={2026},
  booktitle = iclr,
}

@inproceedings{duisterhof2025mast3r,
  title={Mast3r-sfm: a fully-integrated solution for unconstrained structure-from-motion},
  author={Duisterhof, Bardienus Pieter and Zust, Lojze and Weinzaepfel, Philippe and Leroy, Vincent and Cabon, Yohann and Revaud, Jerome},
  booktitle={3DV},
  year={2025},
}

@inproceedings{wang2026vggtomega,
      title={{VGGT-$\Omega$}}, 
      author={Jianyuan Wang and Minghao Chen and Shangzhan Zhang and Nikita Karaev and Johannes L. Sch{\"o}nberger and Patrick Labatut and Piotr Bojanowski and David Novotny and Andrea Vedaldi and Christian Rupprecht},
      booktitle=cvpr,
      year={2026}
}

@inproceedings{wang2025moge2,
      title={MoGe-2: Accurate Monocular Geometry with Metric Scale and Sharp Details}, 
      author={Ruicheng Wang and Sicheng Xu and Yue Dong and Yu Deng and Jianfeng Xiang and Zelong Lv and Guangzhong Sun and Xin Tong and Jiaolong Yang},
      year={2025},
      booktitle=nips
}

@inproceedings{xu2026r33dreconstructionrelative,
      title={$R^3$: 3D Reconstruction via Relative Regression}, 
      author={Congrong Xu and Huachen Gao and Xingyu Chen and Yuliang Xiu and Jun Gao and Anpei Chen},
      year={2026},
      booktitle=nips
}

@inproceedings{burzio2026dejaview,
  title   = {D\'ej\`a View: Looping Transformers for Multi-View 3D Reconstruction},
  author  = {Burzio, Alessandro and Fischer, Tobias and Elflein, Sven and Zhou, Qunjie and de Lutio, Riccardo and Ren, Jiawei and Huang, Jiahui and Huang, Shengyu and Pollefeys, Marc and Leal-Taix{\'e}, Laura and Gojcic, Zan and Turki, Haithem},
  booktitle=nips,
  year    = {2026}
}

@inproceedings{shen2026fastvggt,
  title={FastVGGT: Fast Visual Geometry Transformer},
  author={Shen, You and Zhang, Zhipeng and Qu, Yansong and Zheng, Xiawu and Ji, Jiayi and Zhang, Shengchuan and Cao, Liujuan},
  booktitle=iclr,
  year={2026}
}

@inproceedings{shu2026litevggtboostingvanillavggt,
      title={LiteVGGT: Boosting Vanilla VGGT via Geometry-aware Cached Token Merging}, 
      author={Zhijian Shu and Cheng Lin and Tao Xie and Wei Yin and Ben Li and Zhiyuan Pu and Weize Li and Yao Yao and Xun Cao and Xiaoyang Guo and Xiao-Xiao Long},
      booktitle=cvpr,
      year={2026}
}

@inproceedings{wang2026flashvggtefficientscalablevisual,
      title={FlashVGGT: Efficient and Scalable Visual Geometry Transformers with Compressed Descriptor Attention}, 
      author={Zipeng Wang and Dan Xu},
      booktitle=cvpr,
      year={2026}
}

@inproceedings{wang2026blocksparseglobalattentionefficient,
      title={Block-Sparse Global Attention for Efficient Multi-View Geometry Transformers}, 
      author={Chung-Shien Brian Wang and Christian Schmidt and Jens Piekenbrinck and Bastian Leibe},
      booktitle=cvpr,
      year={2026}
}

@misc{Dellaert26zenodo_GTSAM_4_3,
  howpublished = {Zenodo},
  author       = {Dellaert, Frank and {GTSAM Contributors}},
  title        = {GTSAM 4.3.0},
  month        = sep,
  year         = 2026,
  publisher    = {Zenodo},
  version      = {4.3.0},
  doi          = {10.5281/zenodo.22866773},
  url          = {https://doi.org/10.5281/zenodo.22866773},
}

@article{kaess2012isam2,
  title={iSAM2: Incremental smoothing and mapping using the Bayes tree},
  author={Kaess, Michael and Johannsson, Hordur and Roberts, Richard and Ila, Viorela and Leonard, John J. and Dellaert, Frank},
  journal={The International Journal of Robotics Research},
  volume={31},
  number={2},
  pages={216--235},
  year={2012},
  publisher={Sage Publications Sage UK: London, England}
}

@inproceedings{schops2019bad,
  title={Bad slam: Bundle adjusted direct rgb-d slam},
  author={Sch{\"o}ps, Thomas and Sattler, Torsten and Pollefeys, Marc},
  booktitle=cvpr,
  year={2019}
}

@article{paul2026viewssceneevaluatingmultiview,
  title={Can These Views Be One Scene? Evaluating Multiview 3D Consistency when 3D Foundation Models Hallucinate},
  author={Paul, Soumava and Kaushik, Prakhar and Yuille, Alan},
  journal={arXiv preprint arXiv:2605.18754},
  year={2026},
  html={https://arxiv.org/abs/2605.18754},
  website={https://mvp18.github.io/3d-consistency-metrics/},
  code={https://github.com/mvp18/3DConsistency-metrics}
}

@inproceedings{xiangli2026honey,
  title={{Honey, I Shrunk the Arc de Triomphe!}},
  author={Xiangli, Yuanbo and Chen, Hanyu and Tsang, Xueqing and Snavely, Noah},
  booktitle=eccv,
  year={2026}
}

@inproceedings{shotton2013scene,
  title={Scene coordinate regression forests for camera relocalization in RGB-D images},
  author={Shotton, Jamie and Glocker, Ben and Zach, Christopher and Izadi, Shahram and Criminisi, Antonio and Fitzgibbon, Andrew},
  booktitle=cvpr,
  pages={2930--2937},
  year={2013}
}

@inproceedings{maggio2026vggt,
  title={Vggt-slam: Dense rgb slam optimized on the sl (4) manifold},
  author={Maggio, Dominic and Lim, Hyungtae and Carlone, Luca},
  booktitle=nips,
  year={2025}
}

@article{maggio2026vggt2,
  title={VGGT-SLAM 2.0: Real-time dense feed-forward scene reconstruction},
  author={Maggio, Dominic and Carlone, Luca},
  journal={arXiv preprint arXiv:2601.19887},
  year={2026}
}

@inproceedings{mandal2026vggt,
  title={Vggt-slam++},
  author={Mandal, Avilasha and Kumar, Rajesh and Harithas, Sudarshan Sunil and Arora, Chetan},
  booktitle=cvprw,
  year={2026}
}

@article{yu2026graph3runifyingviewgraph,
      title={Graph3R: Unifying View Graph Initialization with Feed-Forward Reconstruction}, 
      author={Jingwen Yu and Hengli Wang and Jianhao Jiao and Zhonghang Liu and Ping Tan and Hong Zhang},
      journal={arXiv preprint arXiv:2610.16272},
      year={2026},
      eprint={2610.16272},
      archivePrefix={arXiv},
      primaryClass={cs.CV},
      url={https://arxiv.org/abs/2610.16272}, 
}

@inproceedings{kriplani2026archworld,
  title     = {Towards Reconstructing Geographically Diverse Architecture with 3D Foundation Models},
  author    = {Kriplani, Nicky and Zhang, Yiwen and Wang, Angelina and Averbuch-Elor, Hadar},
  booktitle = nips,
  year      = {2026}
}

@inproceedings{olsson2025certifiably,
    author    = {Olsson, Carl and Lochman, Yaroslava and Malmport, Johan and Zach, Christopher},
    title     = {Certifiably Optimal Anisotropic Rotation Averaging},
    booktitle   = iccv,
    year      = {2025},
}

@inproceedings{shah2024flashattention3fastaccurateattention,
      title={FlashAttention-3: Fast and Accurate Attention with Asynchrony and Low-precision}, 
      author={Jay Shah and Ganesh Bikshandi and Ying Zhang and Vijay Thakkar and Pradeep Ramani and Tri Dao},
      year={2024},
      booktitle=nips
}

@inproceedings{lochman2025making,
    author    = {Yaroslava Lochman and Carl Olsson and Christopher Zach},
    title     = {Making Rotation Averaging Fast and Robust with Anisotropic Coordinate Descent},
    booktitle = bmvc,
    year      = {2025},
    url       = {https://bmva-archive.org.uk/bmvc/2025/assets/papers/Paper_516/paper.pdf}
}

@misc{image-matching-challenge-2023,
    author = {Ashley Chow and Eduard Trulls and HCL-Jevster and Kwang Moo Yi and lcmrll and old-ufo and Sohier Dane and tanjigou and WastedCode and Weiwei Sun},
    title = {Image Matching Challenge 2023},
    year = {2023},
    howpublished = {\url{https://www.kaggle.com/competitions/image-matching-challenge-2023}},
    note = {Kaggle}
}

@inproceedings{do2022learning,
  title={Learning to detect scene landmarks for camera localization},
  author={Do, Tien and Miksik, Ondrej and DeGol, Joseph and Park, Hyun Soo and Sinha, Sudipta N.},
  booktitle=cvpr,
  year={2022}
}

@inproceedings{pan2026gluemap,
    author={Pan, Linfei and Sch\"{o}nberger, Johannes Lutz and Pollefeys, Marc},
    title={Global Structure-from-Motion Meets Feedforward Reconstruction},
    booktitle=cvpr,
    year={2026},
}

@inproceedings{durso2026epo,
  title     = {Boosting 3D Foundation Models with Edge-Based Pose Optimization},
  author    = {Mattia D'Urso and Christian Sormann and Mattia Rossi and Friedrich Fraundorfer},
  booktitle = eccv,
  year      = {2026},
}

@misc{Taher_2026_LatentBA,
    author = {Taher, Marwan and Dexheimer, Eric and Davison, Andrew},
    title  = {Latent Bundle Adjustment},
    note   = {Preprint},
    year   = {2026}
}

@inproceedings{ziliotto2026covggt,
  title     = {What VGGT Knows About Overlap: Probing Geometric Foundation Models for Co-Visibility},
  author    = {Ziliotto, Filippo and Serafini, Luciano and Ballan, Lamberto and Campari, Tommaso},
  booktitle = eccv,
  year      = {2026}
}

@article{ORBSLAM3_TRO,
  title={{ORB-SLAM3}: An Accurate Open-Source Library for Visual, Visual-Inertial 
           and Multi-Map {SLAM}},
  author={Campos, Carlos AND Elvira, Richard AND G{\'o}mez Rodr{\'i}guez, Juan J. AND Montiel, 
          Jos\'e M. M. AND Tard\'os, Juan D.},
  journal={IEEE Transactions on Robotics}, 
  volume={37},
  number={6},
  pages={1874-1890},
  year={2021}
 }

@inproceedings{teed2021droid,
  title={{DROID-SLAM: Deep Visual SLAM for Monocular, Stereo, and RGB-D Cameras}},
  author={Teed, Zachary and Deng, Jia},
  booktitle=nips,
  year={2021}
}

@inproceedings{Berton_CVPR_2022_CosPlace,
    author    = {Berton, Gabriele and Masone, Carlo and Caputo, Barbara},
    title     = {Rethinking Visual Geo-Localization for Large-Scale Applications},
    booktitle = cvpr,
    year      = {2022}
}

@InProceedings{Ali-bey_2024_CVPR,
    author    = {Ali-bey, Amar and Chaib-draa, Brahim and Gigu\`ere, Philippe},
    title     = {{BoQ}: A Place is Worth a Bag of Learnable Queries},
    booktitle = cvpr,
    year      = {2024}
}

@article{hubert1985comparing,
  title={Comparing partitions},
  author={Hubert, Lawrence and Arabie, Phipps},
  journal={Journal of Classification},
  volume={2},
  number={1},
  pages={193--218},
  year={1985}
}

@inproceedings{pataki2026vidmap,
  author    = {Zador Pataki and
               Paul-Edouard Sarlin and
               Marc Pollefeys},
  title     = {{VidMap: Exploiting Temporal Structure for Video-Based Structure-from-Motion}},
  booktitle = eccv,
  year      = {2026}
}
}
\appendix
\clearpage
\setcounter{page}{1}
\maketitlesupplementary

\section{Implementation Details}

\paragraph{Training data.} We train on a wide range of 3D vision datasets containing images with ground truth camera parameters. In \cref{tab:dataset_mix}, we mark the datasets that have metric camera poses and those that have reliable ground-truth dense depth maps. Note that there is no overlap between the training scenes and the scenes used for evaluation in \cref{sec:results}.

\paragraph{GT normalization and scale.}\label{suppl:gt-scale} The GT scale $s^*$ is the mean distance, in meters, of the GT camera centers from the first camera, \ie, the factor by which the GT poses are normalized. For mixed-scene sequences, the poses of each scene are normalized separately, and $\mathcal{L}_s$ is not applied, as the scenes share no common scale.

\paragraph{Teacher on mixed-scene inputs.} For multi-scene inputs, we mask the teacher's cross-frame attention across scene boundaries, making its features equivalent to independent per-scene processing while reducing the teacher cost by approximately $2\times$.

\paragraph{Pairwise head at inference.} Since pairs are processed independently, we evaluate the $\mathcal{O}(N^2)$ pairs input to $h_\mathrm{rel}$ (\cref{subsec:relative-pose}) in blocks at inference, which makes the head's peak memory constant in $N$.

\paragraph{Baselines and timing.} VGGT-$\Omega_{512}$ and VGGT-$\Omega_{256}$ are two separate VGGT-$\Omega$ checkpoints, trained at $512$ and $256$ pixel resolution, respectively, and each is evaluated at the resolution it was trained for. VGGT-$\Omega_{256}$ is also the teacher used for distillation. All reported runtimes of \ours~include pose averaging, \ie, they cover the full pipeline from images to absolute poses. Pose averaging computes all per-edge quantities in batch on the GPU, while its inherently sequential steps, \ie, the coordinate-descent sweeps of rotation averaging~\cite{lochman2025making} and the sequential placement in chaining, run on the CPU. Compared to a direct GPU implementation, this reduces anisotropic averaging of 1,500 frames from 6.2\,s to 0.65\,s with identical outputs.

\paragraph{Architecture ablation.} In \cref{tab:arch-ablation}, all rows share a trunk of 12 frame attention and 12 global attention blocks (hidden dimension 384), without weight sharing until the second-to-last row. Rows without relative pose replace $h_\mathrm{rel}$ with a VGGT-style per-frame camera head that predicts absolute poses relative to the first frame, trained with an L1 loss on the pose encoding. The first row trains this head on the GT (normalized as in \cref{subsec:loss}) without distillation. The two distillation rows instead use the absolute poses predicted by the VGGT-$\Omega_{256}$ teacher as targets, together with feature distillation from the teacher's last layer or from every level; \emph{+ GT supervision} switches the targets back to the GT. Relative pose trades a small loss in AUC@$30^\circ$ for higher MegaDepth AUC@$3^\circ$, and enables $\hat{c}_{ij}$ and $\hat{\boldsymbol{\Lambda}}_{ij}$.

\paragraph{Conditioning on known quantities.} We condition half of the training sequences, restricting conditioning to single-scene sequences. For each conditioned sequence, we randomly provide $20$--$80\%$ of the poses, and independently provide focal lengths and metric scale. The provided values are not perturbed. The camera loss is applied to all edges, including known edges, which are additionally trained toward maximal confidence. This encourages the predicted edges to remain consistent with the supplied measurements. Pose targets are expressed in the normalized unit $\bar{s}$, and feature distillation is applied only to unconditioned sequences.

\begin{table}[t] \centering
\small 
\caption{\textbf{Dataset mixture for \ours.} \emph{Metric} marks datasets whose camera poses are in meters, \emph{Depth} those that ship dense ground-truth depth maps. We train and evaluate on disjoint scenes.}
\label{tab:dataset_mix}
\setlength{\tabcolsep}{4pt}
\begin{tabular}{lccc}
\toprule
\textbf{Dataset} & \textbf{Type} / \textbf{Source} & \textbf{Metric} & \textbf{Depth} \\
\midrule
AerialMD~\citep{vuong2025aerialmegadepth}        & Aerial / MVS           & \cmark & \cmark \\
ASE~\citep{AriaSynthEnv:2025}             & Indoor / Graphics      & \cmark & \cmark \\
ArchWorld~\citep{kriplani2026archworld}             & Outdoor / MVS      & \cmark & \cmark \\
BlendedMVS~\citep{yao2020blendedmvs}             & Aerial / Mesh          & \xmark & \cmark \\
DL3DV~\citep{ling2024dl3dv}                      & Mixed / Video          & \xmark & \xmark \\
Hypersim~\citep{roberts2021hypersim}             & Indoor / Graphics      & \cmark & \cmark \\
Map-Free~\citep{arnold2022map}                   & Object / MVS   & \cmark & \xmark \\
MegaDepth~\citep{li2018megadepth}                & Outdoor / MVS          & \xmark & \cmark \\
MegaDepth-X~\citep{li2026longtail}                   & Outdoor / MVS          & \xmark & \cmark \\
MegaScenes~\citep{tung2024megascenes}            & Outdoor / SfM          & \xmark & \xmark \\
MegaSynth~\citep{Jiang_2025_CVPR}                      & Procedural / Graphics  & \xmark & \cmark \\
MetricScenes~\citep{xiangli2026honey}                & Outdoor / MVS          & \cmark & \cmark \\
MPSD~\citep{mpsd:2020}                                & Outdoor / MVS          & \xmark & \cmark \\
RealEstate10K~\citep{realestate10k}              & Indoor / Video         & \xmark & \xmark \\
ScanNet++~\citep{yeshwanth2023scannet++}         & Indoor / Mesh          & \cmark & \cmark \\
SpatialVID~\citep{wang2025spatialvid}            & Outdoor / Video        & \xmark & \xmark \\
TerraSky3D~\citep{durso202Xterrasky3d}                    & Outdoor / MVS          & \xmark & \cmark \\
VKITTI 2~\citep{gaidon2016virtual,cabon2020vkitti2} & Outdoor / Graphics & \cmark & \cmark \\
\bottomrule
\end{tabular}
\end{table}

\subsection{Details of the Visual Localization Experiments}\label{suppl:visloc}
\paragraph{Multiple-reference experimental setup.}
On Cambridge Landmarks~\cite{kendall2015posenet} and Indoor6~\cite{do2022learning}, we follow Reloc3r~\cite{reloc3r:2025} and retrieve the top-$10$ reference images per query using NetVLAD~\cite{arandjelovic2016netvlad}. We consider calibrated and uncalibrated settings, depending on whether the query intrinsics are known. In both settings, \ours$_{\mathbf{T}}$ conditions on the known reference poses ($\delta^{\mathbf{T}}_i$ in \cref{subsec:conditioning}). In the calibrated setting, it additionally receives the focal lengths of all input images, including the query ($\delta^{\mathrm{f}}_i$, see \cref{eq:conditioning}). We also evaluate pose-conditioned variants of $\pi^3$ and MapAnything in the corresponding settings. DA3-G's released implementation requires poses for all input images when pose conditioning is enabled, so we do not evaluate it with reference-only pose conditioning.

For VGGT-$\Omega$, DA3-G, and \dejaview{} that predict camera poses in a shared coordinate system, we estimate a $\mathrm{Sim}(3)$ transformation by Umeyama alignment~\cite{umeyama1991leastsquares} of the predicted reference camera centers to their known counterparts. Then, we apply it to the predicted query pose. For Reloc3r and $R^3$, which predict pairwise relative poses, we use Reloc3r's motion averaging to recover the query pose in the coordinate system of the known references. Since \ours$_\mathbf{T}$, $\pi^3_\mathbf{T}$ and MapAny.$_{\mathbf{T}}$ condition on the known reference poses, they directly predict the query pose in the same coordinate system, requiring no further alignment.
All experiments are run on an NVIDIA GH200.

\paragraph{Single-reference experimental setup.}
Single-reference localization estimates the metric relative pose between a query and a reference image. For \ours, $\pi^3$, and $R^3$, we use their metric-scale pose predictions directly.

For VGGT-$\Omega$ and \dejaview{}, which predict point maps without metric scale, we precompute and cache metric reference depth using DA3-G. For each query--reference pair, we align the predicted reference point map to this depth to recover the scale of the relative translation. This preprocessing is performed once per scene, taking $73.0$~ms per reference image with $7.3$~GiB peak GPU memory usage. These preprocessing costs are excluded from the reported per-query runtime and peak GPU memory.

Map-Free also requires a confidence score for each query to rank pose estimates for AUC evaluation~\cite{arnold2022map}. AUC integrates localization precision over the fraction of queries retained as the confidence threshold varies. For VGGT-$\Omega$, \dejaview{}, DA3-G, $\pi^3$, and MapAnything, which do not predict pose confidence, we average each model's own per-pixel depth or point-map confidence over the query image. $R^3$ uses its predicted mean translation and rotation confidence as mentioned in their paper.

For \ours, we derive confidence from the predicted $6\times6$ pose information matrix $\hat{\boldsymbol{\Lambda}}_{ij}$ between query and reference. Marginalizing out translation (as in \cref{sec:pose-avr}) gives the rotation information matrix
\begin{equation}
\boldsymbol{\Lambda}^{\omega}_{ij}
    = \hat{\boldsymbol{\Lambda}}_{ij}^{\omega\omega}
    - \hat{\boldsymbol{\Lambda}}_{ij}^{\omega t}\left({\hat{\boldsymbol{\Lambda}}_{ij}^{tt}}\right)^{-1}
      \hat{\boldsymbol{\Lambda}}_{ij}^{t \omega},
    \quad
    q = |\boldsymbol{\Lambda}^{\omega}_{ij}|^{1/3},
\end{equation}
where the superscripts $\omega$ and $t$ index the rotation and translation blocks of $\hat{\boldsymbol{\Lambda}}_{ij}$, respectively. We rank query poses by $q$, the geometric mean of the eigenvalues of $\boldsymbol{\Lambda}^{\omega}_{ij}$. This same matrix weights the rotation residuals in our anisotropic pose averaging.

We use rotation information because the edge translations are predicted in normalized units, while metric scale is predicted by a separate head whose errors are not modeled by $\hat{\boldsymbol{\Lambda}}_{ij}$. We prefer $q$ over the predicted scalar confidence because the latter is trained by regression to the bounded target $1-(\theta_R+\theta_t)/(2\pi)$ and tends to saturate near one, limiting its ability to distinguish accurate predictions. 
Instead, we use the rotational precision from the information matrix learned through a Gaussian negative log-likelihood.
For the benchmark submission, we use the normalized ascending rank of $q$ across all $M$ test queries as the confidence score.
Results are shown in~\cref{suppl:single-ref-visloc}.

\subsection{Details of the SfM Experiment}\label{suppl:sfm}
We provide implementation and evaluation details for the SfM experiment in \cref{subsec:results:sfm}.

\paragraph{Initialization.}
We run \ours~once over all images of a scene, resized
to a long side of 256 pixels with the aspect ratio preserved and padded to a
common $256\times256$ canvas. Absolute poses are obtained from the predicted relative poses with our anisotropic pose averaging, keeping edges with confidence at least 0.9. Each image gets its own camera (simple radial model) with the predicted focal length $((\hat f_x + \hat f_y)/2$ for \ours$)$, the principal point at the image center, and zero distortion.

\paragraph{Matches.}
We use the database from GLOMAP's run: SIFT features, exhaustive matching, and two-view verification with COLMAP's settings for its global mapper (1\,px maximum Sampson error, at least 30 inliers, inlier ratio at least 0.25).

\paragraph{Triangulation.}
Under the predicted poses, we triangulate 3D points from the verified matches with COLMAP's incremental triangulator (2\textdegree\ maximum angular error for creating and continuing tracks, 1\textdegree\ minimum triangulation angle).

\paragraph{Refinement.}
We use the iterative global refinement procedure that COLMAP’s incremental mapper runs after image registration. The procedure starts with track completion and merging, retriangulation of under-reconstructed image pairs, then passes of global bundle adjustment, track completion and merging, and outlier filtering. We use two passes, a Huber loss (1\,px) in the bundle adjustment, a 15\,px reprojection gate for completing and merging tracks (default 4\,px), and exclude tracks shorter than three observations from the bundle adjustment. We finish with a least-squares bundle adjustment (at most 200 iterations) on this inlier set. Each bundle adjustment refines camera poses, 3D points, focal lengths, and radial distortion while keeping the principal points fixed. We fix the similarity gauge by holding one camera pose and one translation component of a second camera constant.

\paragraph{Timing.}
All methods run on the same machine (NVIDIA RTX 4090, 32 CPU
cores). The baselines use COLMAP 4.1.1 (\texttt{automatic\_reconstructor}, 64
threads, GPU enabled, default settings). Our timing covers model loading,
inference, and the refinement, which runs on the CPU. We report the median of
three runs (COLMAP: a single run due to its long runtime) and the mean AUC over the scenes of each run.

\paragraph{Evaluation.}
For every pair of ground-truth images, the error is the larger of the relative rotation error and the translation-direction error.
Pairs whose ground-truth baseline is below 2\,cm contribute no translation
error. An image that is not registered fails every pair it is in.

\subsection{Details of the Visual Odometry Experiment}\label{suppl:vo}
Here, we provide implementation and evaluation details for the visual
odometry experiments in \cref{subsec:results:vo}.

\paragraph{Keyframes and blocks.}
A frame becomes a keyframe when the maximum cosine similarity $\psi_i$ between its DINOv3 class token~\cite{simeoni2025dinov3} (which is already computed in the \ours~forward pass) and those of the keyframes it would share a block with lies in $(\psi_{\mathrm{lo}}, \psi_{\mathrm{hi}}) = (0.85, 0.95)$: low enough that it adds new content, high enough to retain visual overlap. If no frame qualifies within twice the median keyframe spacing, the next frame is accepted anyway ($\psi_i \leq \psi_\mathrm{lo}$ must hold as well). Each time $\Delta=4$ new keyframes have been selected, the $B=16$ most recent ones form a block $\mathcal{B}_j$, which we process in one joint forward pass.

\paragraph{Pose graph.}
Each block gets a scale variable $\beta_j$. Every confident pair $(i,k)$ in the block adds a constraint between the global keyframe poses $\mathbf{T}_i, \mathbf{T}_k \in \mathrm{SE}(3)$ and $\beta_j$. Consecutive blocks share $B-\Delta$ keyframes, which chain them together, and a new block is initialized by a RANSAC--Umeyama~\cite{umeyama1991leastsquares} $\mathrm{Sim}(3)$ fit on these shared keyframes. After each block, we update the graph incrementally with iSAM2~\cite{kaess2012isam2} in GTSAM~\cite{Dellaert26zenodo_GTSAM_4_3}.

\paragraph{Tracking.}
Every frame is localized in a single pass against the most recent block. The frame's tokens attend to the block's cached keys and values, but the block does not attend back, which reduces the attention cost from $\mathcal{O}(B^2P^2)$ to $\mathcal{O}(BP^2)$ for $P$ tokens per frame. Each confident edge to a block keyframe gives a candidate pose, and we fuse the candidates with an information-weighted mean. Non-keyframes stay out of the graph. Instead, their poses are recomputed from the stored edges whenever optimization moves their reference keyframes or rescales $\beta_j$.

\paragraph{Global passes.}
Every $G=4$ blocks, we run a joint pass over keyframes from distant parts of the trajectory to obtain loop closures.
For each keyframe added since the last pass, we retrieve up to \(B-1\) keyframes most similar to it outside all blocks containing it, retaining
only those with similarity $\geq \psi_{\mathrm{lo}}$. We select as anchor
the keyframe with the largest summed similarity to its retrieved
keyframes, and process it together with them.
As with a block, the pass gets its own scale variable, and its confident edges are added to the graph.

\begin{table}[t]
\centering
\caption{\textbf{Final trajectory pose coverage (\%).}
Percentage of input frames represented in the final output trajectory,
pooled over all sequences, including failed sequences. Low-resolution inputs are
$144\times256$ for Cambridge and $160\times256$ for ETH3D ($H\times W$).}
\label{tab:pose_coverage}
\small
\setlength{\tabcolsep}{2.8pt}
\begin{tabular}{l rrrr}
\toprule
& \multicolumn{2}{c}{\textbf{Native input}}
& \multicolumn{2}{c}{\textbf{Low-resolution input}} \\
\cmidrule(lr){2-3}
\cmidrule(lr){4-5}
\textbf{Method} & Cambridge & ETH3D & Cambridge & ETH3D \\
\midrule
DROID-SLAM    & \best 100.0 & \best 100.0 & \best 100.0 & \best 100.0 \\
ORB-SLAM3     &   7.1 &  \good 27.8 &  \good 8.5 &  \good 9.4 \\
VGGT-SLAM 2.0 &  \good 58.8 &   7.5 &   1.1 &   2.3 \\
\ours~(ours)  & \best 100.0 & \best 100.0 & \best 100.0 & \best 100.0 \\
\bottomrule
\end{tabular}
\end{table}

\subsubsection{Evaluation Details}
FPS is the total number of input frames divided by the total processing time (including initialization and final trajectory completion). Post-initialization first-pose latency is the mean time from a frame's input to its first output pose, measured over frames that receive an online pose after initialization. These poses may subsequently be refined by, \eg, loop closure. Initialization ends when the method outputs its first pose. Time spent resizing images to each model's native resolution is excluded from both FPS and latency measurements. Accuracy is evaluated on the final output trajectories. VRAM is the peak PyTorch-allocated GPU memory over all sequences. For absolute trajectory error (ATE), we align each trajectory to the ground truth with a least-squares $\mathrm{Sim}(3)$~\cite{umeyama1991leastsquares} and average the RMSE of the camera centers over sequences. To separate local accuracy from outliers, we also report the median translation error $t$ and rotation error $R$ after a RANSAC-based $\mathrm{Sim}(3)$ alignment.

\paragraph{Baseline pose-output protocols.}
ORB-SLAM3 processes every input frame and exposes online poses when tracking succeeds. Initialization and tracking loss can reduce pose coverage. For DROID-SLAM, online pose latency is measured after initialization on frames accepted by its image-motion-based frame selector. Poses for the remaining frames are estimated during final trajectory completion. VGGT-SLAM 2.0 estimates poses only for disparity-selected frames and releases them in submap batches. Its FPS therefore measures input-frame throughput, while pose latency is evaluated only on frames receiving an online estimate. Latency is therefore evaluated on method-dependent subsets of input frames. Final pose coverage also includes poses produced during trajectory completion.

\paragraph{Final pose coverage.}
Pose coverage is the percentage of input frames represented in the final output trajectory, aggregated over all sequences. The pose coverage is reported in~\cref{tab:pose_coverage}.

\section{Additional Experiments}
\begin{figure}[t]
\centering
\includegraphics[width=0.99\linewidth]{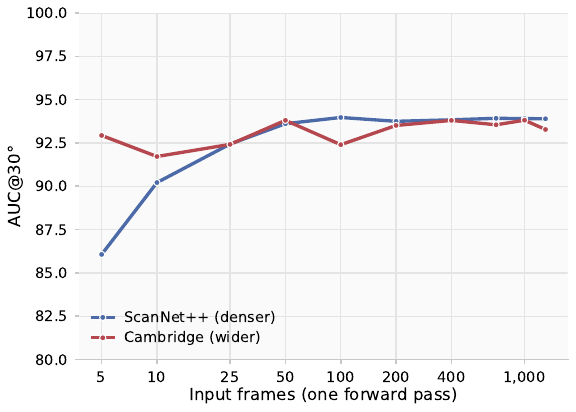}
\caption{\textbf{Pose accuracy vs.\ number of input frames.} AUC@30 of \ours{} on ScanNet++ and Cambridge Landmarks, single forward pass. Accuracy stays flat up to 1,300 frames.}
\label{fig:auc-vs-frames}
\end{figure}

\subsection{Pose Accuracy for Long Sequences}\label{suppl:long}

\begin{figure*}[t]
\centering
\includegraphics[width=0.99\linewidth]{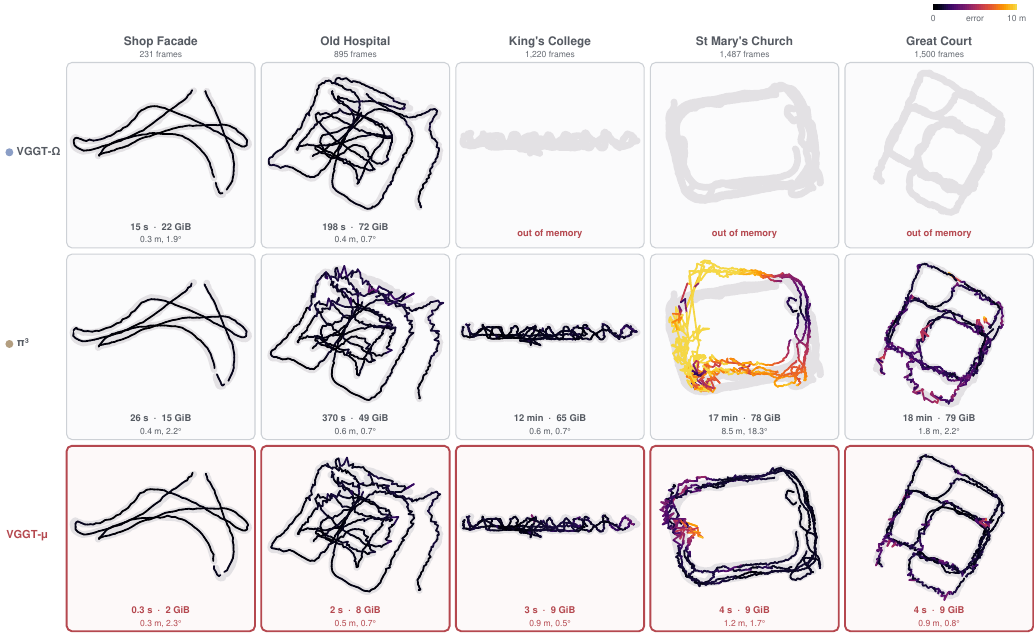}
\caption{\textbf{Qualitative comparison on long sequences.} All mapping frames of each Cambridge Landmarks scene in one forward pass (protocol in \cref{suppl:long}). Trajectories are aligned to the ground truth (gray) by a similarity transform and colored by their camera-center error. Each cell gives runtime, peak memory, and mean position and rotation error. VGGT-$\Omega$ runs out of memory on the three scenes with more than 1,000 frames (1,220--1,500), and $\pi^3$ needs up to 18\,min and 79\,GiB, while \ours~needs at most 4\,s and 9\,GiB.}
\label{fig:qualitative-cambridge}
\end{figure*}

\paragraph{Protocol.} For \cref{fig:teaser}(a), \cref{fig:qualitative} (1,000 frames sampled evenly over the capture), and \cref{fig:qualitative-cambridge}, we use the mapping (training) split of the Cambridge Landmarks~\cite{kendall2015posenet} scenes, kept in capture order: all frames of each scene, except for Great Court, where we sample 1,500 of its 1,531 frames evenly over the capture. Each model receives all frames in a single forward pass, on one NVIDIA GH200 (96\,GiB), at its native resolution: long side 256 for \ours, and a pixel budget of $512^2$ for VGGT-$\Omega$ and $504^2$ for $\pi^3$ (aspect ratio preserved). An out-of-memory error counts as a failure. For $\pi^3$ on Great Court, we set PyTorch's \texttt{expandable\_segments} allocator option, without which it fails due to memory fragmentation. Runtime is the wall time of the forward pass including pose averaging, and memory is the peak allocated GPU memory, weights included. Cambridge frames are $854\times480$, which \ours~processes at $256\times144$, \ie, $16\times9=144$ tokens per frame. The efficiency comparisons in \cref{tab:runtime,tab:efficient-vggt} instead use square $256\times256$ inputs with 256 tokens per frame, so the same number of frames costs more there: 1,000 Cambridge frames take 2.2\,s, whereas 1,000 square frames take 4.6\,s, more than the 1,500 Cambridge frames of \cref{fig:teaser}(a). To compute errors, we align the estimated camera centers to the ground truth with a similarity transform~\cite{umeyama1991leastsquares} and report the mean camera-center error (in meters; the ground-truth reconstructions are approximately metric) and the mean rotation error.

\paragraph{Accuracy vs.\ sequence length.} \Cref{fig:auc-vs-frames} evaluates multi-view relative pose (as in \cref{tab:mvrelpose}) on ScanNet++ and on the Cambridge Landmarks mapping sequences, with 5 to 1,300 frames per sequence in a single forward pass.

\paragraph{Qualitative results.} \Cref{fig:qualitative-cambridge} shows the trajectories of all five scenes. VGGT-$\Omega$ runs out of memory on the three longest scenes. \ours~is about as accurate as the baselines on the shorter scenes and clearly more accurate than $\pi^3$ on St Mary's Church and Great Court, while running 90--260$\times$ faster than $\pi^3$.

\subsection{Comparison with Efficient VGGT Variants}\label{suppl:efficient-vggt}
FastVGGT~\cite{shen2026fastvggt} and LiteVGGT~\cite{shu2026litevggtboostingvanillavggt} speed up VGGT on long sequences by merging tokens in the global attention, LiteVGGT additionally runs in FP8 after fine-tuning. \Cref{tab:efficient-vggt} compares them, and VGGT-$\Omega_{512}$, with \ours~on 100 to 1,000 frames (\cref{fig:pareto-long} in the main paper shows Cambridge Landmarks at 1,000 frames). At 1,000 frames, \ours~is $20\times$ faster than LiteVGGT and $44\times$ faster than VGGT-$\Omega$, the most accurate baseline, with $7$--$8\times$ less memory, and at least 12 AUC@$3^\circ$ points more accurate on both datasets. Since token merging is orthogonal to our design, we also apply FastVGGT's training-free merging to the global attention of \ours~and of VGGT-$\Omega$: before every global attention, all tokens of the first frame, the camera and register tokens, one patch per $2\times2$ cell of the patch grid, and every tenth token are kept, and every other token is merged into its most similar kept patch token. On 1,000 frames, this speeds up both models by $1.9\times$, for at most 1 AUC@$30^\circ$ point for \ours~from 500 frames on. On 100 frames, where the global attention is a small part of the runtime, merging saves little time and costs more accuracy. The merge assignment has to be recomputed before every iteration of the recurrent block: computed once and reused, it fails.

\begin{table}[t]
\centering
\caption{\textbf{Comparison with efficient VGGT variants.} Multi-view relative pose in a single forward pass, and runtime (s) and peak GPU memory (GiB) on one GH200. ``+ merging'': training-free token merging of FastVGGT applied to the model above.}
\label{tab:efficient-vggt}
\setlength{\tabcolsep}{2.4pt}
\begin{tabular}{l cc cc rr}
\toprule
& \multicolumn{2}{c}{\textbf{ScanNet++}} & \multicolumn{2}{c}{\textbf{Cambridge}} & & \\
\cmidrule(lr){2-3}\cmidrule(lr){4-5}
\textbf{Method} & @$3^\circ$ & @$30^\circ$ & @$3^\circ$ & @$30^\circ$ & Time & Mem. \\
\midrule
\multicolumn{7}{l}{\textit{100 frames}} \\
FastVGGT & 12.4 & 76.1 & 33.8 & 87.1 & 4.53 & 10.5 \\
LiteVGGT & 21.5 & 80.6 & 36.1 & 86.3 & 2.08 & 9.2 \\
VGGT-$\Omega_{512}$ & 32.7 & 87.3 & 57.4 & \textbf{92.6} & 3.01 & 11.7 \\
\quad + merging & 35.0 & 87.4 & 49.8 & 91.1 & 2.36 & 11.7 \\
\ours & \textbf{65.5} & \textbf{94.0} & \textbf{57.9} & 92.4 & 0.13 & \textbf{1.4} \\
\quad + merging & 58.1 & 92.4 & 47.2 & 89.7 & \textbf{0.12} & \textbf{1.4} \\
\midrule
\multicolumn{7}{l}{\textit{500 frames}} \\
FastVGGT & 12.1 & 75.6 & 28.1 & 84.8 & 69.6 & 44.2 \\
LiteVGGT & 16.1 & 75.4 & 31.9 & 77.6 & 26.5 & 36.1 \\
VGGT-$\Omega_{512}$ & 32.9 & 87.5 & 51.6 & 87.0 & 51.7 & 41.8 \\
\quad + merging & 37.6 & 88.4 & 47.3 & 86.4 & 29.7 & 41.8 \\
\ours & \textbf{67.0} & \textbf{93.9} & \textbf{60.3} & \textbf{93.8} & 1.38 & \textbf{5.3} \\
\quad + merging & 64.9 & 93.8 & 56.9 & 92.8 & \textbf{0.86} & \textbf{5.3} \\
\midrule
\multicolumn{7}{l}{\textit{1,000 frames}} \\
FastVGGT & 11.8 & 74.5 & 22.7 & 73.0 & 246.1 & 86.3 \\
LiteVGGT & 14.7 & 73.3 & 28.3 & 72.3 & 91.7 & 66.5 \\
VGGT-$\Omega_{512}$ & 32.6 & 87.4 & 47.6 & 81.6 & 203.6 & 79.4 \\ %
\quad + merging & 37.6 & 88.5 & 44.1 & 81.1 & 105.1 & 79.4 \\
\ours & \textbf{66.8} & \textbf{93.9} & \textbf{60.1} & \textbf{93.8} & 4.61 & \textbf{9.8} \\
\quad + merging & 64.9 & 93.8 & 58.7 & 93.4 & \textbf{2.47} & \textbf{9.8} \\
\bottomrule
\end{tabular}
\end{table}

\subsection{Training Resolution at Equal Compute}\label{suppl:iso-compute}
\begin{figure}[t]
\centering
\includegraphics[width=0.99\linewidth]{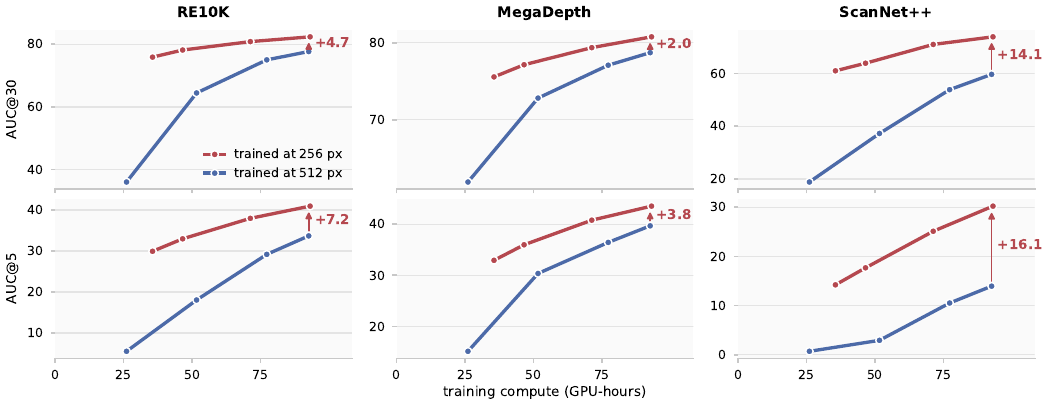}
\caption{\textbf{Training resolution at equal compute.} AUC@$30^\circ$ (top) and AUC@$5^\circ$ (bottom) of checkpoints trained at 256 and 512 pixels, against training GPU-hours, each tested at its training resolution. Arrows mark the gap at the largest budget.}
\label{fig:iso-compute}
\end{figure}

We train \ours~at 256 pixels. To test whether a higher resolution would be a better use of compute, we train the same recipe from scratch at 512 pixels, distilling from VGGT-$\Omega_{512}$ and using gradient checkpointing to fit the same sequence length, and compare checkpoints at equal training GPU-hours (4 GPUs each, evaluations included). Each checkpoint is evaluated on 10-frame sequences at its training resolution. As a training step at 512 pixels costs about $3.7\times$ more, training at 256 pixels is more accurate at every budget (\cref{fig:iso-compute}), most clearly on ScanNet++.

\subsection{Metric Pose Estimation}\label{subsec:results:metric}

\begin{figure}[t]
\centering
\includegraphics[width=0.99\linewidth]{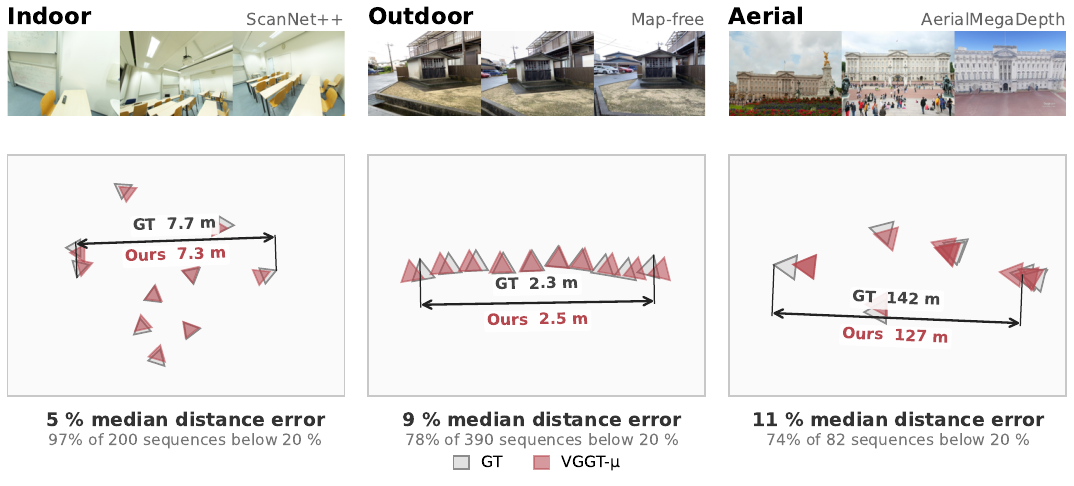}
\caption{\textbf{Metric pose estimates.} Top view of the GT cameras (gray) and the metric cameras of \ours~(red) after a rigid alignment, with the distance between the two farthest-apart cameras as measured by each. Each panel shows the sequence with the median error of its dataset. Below are the statistics over all sequences.} 
\label{fig:metric}
\end{figure}

In \cref{tab:metric}, we compare the accuracy of our metric pose estimates to baselines on Map-Free. \ours~is far more accurate than $\pi^3$-X and MapAnything, and close to DA3-Nested, which combines DA3-Giant with a separate metric depth model (1.7B parameters, $14\times$ as many as \ours). The strong metric pose accuracy on Map-Free allows us to achieve good scores on the Map-Free relocalization benchmark (\cref{tab:mapfree_native_full}) without the help of any additional model. \Cref{fig:metric} visualizes metric cameras indoors, outdoors, and from aerial imagery.

\begin{table}[t]
\centering
\caption{\textbf{Metric camera poses on Map-Free.} Ten-frame sequences. We report the share (\%) of frame pairs whose relative pose is within 25\,cm and $5^\circ$, the share (\%) of sequences whose best-fit scale is within 10\,\% and 25\,\% of the true scale, the median relative translation ($t$) and rotation ($R$) errors over frame pairs, and the median absolute trajectory error (ATE) after a rigid alignment without scaling.}
\label{tab:metric}
\footnotesize
\setlength{\tabcolsep}{3pt}
\begin{tabular}{l rrr rrr}
\toprule
& \textbf{Pairs} $\uparrow$ & \multicolumn{2}{c}{\textbf{Scale} $\uparrow$} & \multicolumn{3}{c}{\textbf{Median error} $\downarrow$} \\
\cmidrule(lr){2-2} \cmidrule(lr){3-4} \cmidrule(lr){5-7}
\textbf{Method} & 25\,cm, $5^\circ$ & $\leq$10\% & $\leq$25\% & $t$ (cm) & $R$ ($^\circ$) & ATE (cm) \\
\midrule
$\pi^3$-X & 86.1 & 20.4 & 46.9 & 6.6 & \better 0.23 & 6.6 \\
MapAnything & 77.2 & 15.4 & 35.4 & 11.3 & 0.49 & 10.7 \\
DA3-Nested & \best 97.4 & \best 54.6 & \best 90.8 & \best 2.6 & \best 0.21 & \best 2.5 \\
\ours & \better 96.7 & \better 46.5 & \better 83.5 & \better 3.7 & 0.28 & \better 3.6 \\
\bottomrule
\end{tabular}
\end{table}

\subsection{Single Reference Visual Localization}\label{suppl:single-ref-visloc}
\begin{figure}[t]
\centering
\includegraphics[width=0.99\linewidth]{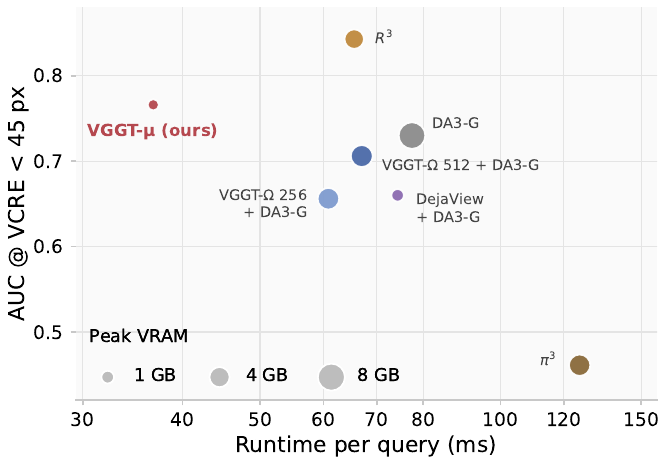}
\caption{\textbf{Single-reference localization on Map-Free.} AUC (VCRE $<45$ pixels) versus runtime per query. Marker size indicates peak GPU memory usage. VGGT-$\Omega$ and \dejaview{} use an additional DA3-G pass on the reference image to recover metric scale. This auxiliary DA3-G inference is excluded from the reported runtime and peak GPU memory.}
\label{fig:mapfree-pareto}
\end{figure}

We then consider localization from only one reference image on Map-Free~\cite{arnold2022map}. A key challenge is accurately estimating the metric scale of translation. While VGGT-$\Omega$ and \dejaview{} use an additional DA3-G pass on the reference image for scale recovery, \ours~directly predicts metric relative poses without auxiliary depth estimation. These predictions enable accurate localization with low runtime and peak GPU memory usage, as shown in \cref{fig:mapfree-pareto}.
Detailed results are given in \cref{tab:mapfree_native_full}, and the scale-recovery procedure is described in \cref{suppl:visloc}.

\begin{figure}[t]
\centering
\includegraphics[width=0.99\linewidth]{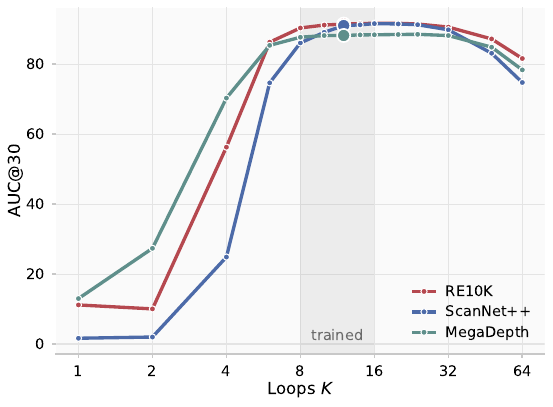}
\caption{\textbf{Number of recurrent iterations.} AUC@$30^\circ$ on 10-frame sequences against the number of iterations $K$ at inference. The shaded band is the range of $K$ sampled during training. The large markers show $K=12$, used throughout the paper.} 
\label{fig:loop-curve}
\end{figure}

\begin{table*}[t]
    \centering
    \caption{\textbf{Visual localization on Map-Free.} Test split, scores from the benchmark server; every model at its native input (\ours: long side 256). Top: images only; bottom (subscript $\mathrm{f}$): both frames' intrinsics given as input. VGGT-$\Omega$ and \dejaview\ take metric scale from DA3-G's depth of the reference. This DA3-G pass (73.0\,ms, 7.3\,GiB per reference image) is excluded from their Time and VRAM. Time: median per query on one GH200, batch 1; VRAM: peak, weights included. Colors indicate ranks within each group.}
    \label{tab:mapfree_native_full}
    \renewcommand{\arraystretch}{1.15}
    \setlength{\tabcolsep}{4pt}
    \resizebox{\linewidth}{!}{%
    \begin{tabular}{@{}lrrrrrrrrrrr@{}}
        \toprule[1pt]
        \textbf{Method} & \multicolumn{2}{c}{\textbf{VCRE $<45$ px}} & \multicolumn{2}{c}{\textbf{VCRE $<90$ px}} & \textbf{VCRE} & \multicolumn{2}{c}{\textbf{25 cm, $5^\circ$}} & \multicolumn{2}{c}{\textbf{Median}} & \textbf{Time} & \textbf{VRAM} \\
        \cmidrule(lr){2-3}\cmidrule(lr){4-5}\cmidrule(lr){7-8}\cmidrule(lr){9-10}
        & AUC $\uparrow$ & Prec. $\uparrow$ & AUC $\uparrow$ & Prec. $\uparrow$ & median (px) $\downarrow$ & AUC $\uparrow$ & Prec. $\uparrow$ & $t$ (m) $\downarrow$ & $R$ ($^\circ$) $\downarrow$ & (ms) $\downarrow$ & (GiB) $\downarrow$ \\
        \midrule
        VGGT-$\Omega_{512}$+DA3-G & 0.706 & \good{56.7\%} & 0.870 & \better{75.4\%} & 71.4 & 0.668 & \good{48.9\%} & 0.54 & 6.3 & 66.4 & 4.93 \\
        VGGT-$\Omega_{256}$+DA3-G & 0.656 & 53.3\% & 0.833 & \good{73.7\%} & 72.2 & 0.602 & 45.2\% & 0.58 & 6.5 & \better{60.2} & 4.73 \\
        \dejaview{}+DA3-G & 0.660 & 52.8\% & 0.846 & 73.1\% & \better{61.6} & 0.524 & 35.2\% & \better{0.49} & \good{5.2} & 73.7 & \better{0.95} \\
        DA3-G & \good{0.730} & \better{58.4\%} & \good{0.894} & \good{73.7\%} & 83.0 & \good{0.686} & \best{51.1\%} & 0.68 & 5.5 & 77.5 & 7.69 \\
        $R^3$ & \best{0.842} & \best{60.3\%} & \best{0.948} & \best{79.1\%} & \best{55.1} & \best{0.790} & \better{50.9\%} & \best{0.45} & \best{4.6} & \good{65.6} & \good{3.65} \\
        $\pi^3$ & 0.461 & 34.1\% & 0.691 & 56.5\% & 112.5 & 0.423 & 28.7\% & 0.81 & \better{5.1} & 125.6 & 4.61 \\
        MapAnything & 0.281 & 23.8\% & 0.534 & 46.4\% & 116.4 & 0.227 & 18.1\% & 1.15 & 6.3 & 196.1 & 7.34 \\
        \ours~(ours) & \better{0.766} & 55.6\% & \better{0.900} & 73.6\% & \good{65.7} & \better{0.695} & 45.3\% & \good{0.53} & 5.9 & \best{36.7} & \best{0.61} \\
        \midrule
        $\pi^3_{\mathrm{f}}$ & \better{0.440} & \better{33.1\%} & \better{0.684} & \better{55.7\%} & \good{113.6} & \better{0.403} & \better{28.2\%} & \better{0.80} & \better{5.1} & \better{135.5} & \better{6.08} \\
        MapAnything$_{\mathrm{f}}$ & \good{0.310} & \good{27.1\%} & \good{0.593} & \good{51.8\%} & \better{107.2} & \good{0.265} & \good{21.4\%} & \good{1.02} & \good{5.7} & \good{195.3} & \good{7.35} \\
        \ours$_{\mathrm{f}}$ & \best{0.755} & \best{54.9\%} & \best{0.889} & \best{72.7\%} & \best{66.4} & \best{0.711} & \best{47.9\%} & \best{0.51} & \best{4.9} & \best{43.0} & \best{0.61} \\
        \bottomrule[1pt]
    \end{tabular}%
    }
\end{table*}

\subsection{Additional Visual Odometry Experiments}\label{suppl:vo-low-res}
The main paper reports the original evaluation runs. Here we provide additional repetitions of each method using the same protocol, reporting medians of three per-run dataset aggregates (mean for peak VRAM). \Cref{tab:odometry-both} also runs the baselines on the same low-resolution input images as we use for \ours~and simultaneously extends \cref{tab:odometry} with additional metrics. 

Tracking outcomes, pose coverage, and timing can vary across runs. In particular, ORB-SLAM3’s accuracy is conditional on the sequences and poses that remain evaluable and thus fluctuates a bit across runs. Success and coverage should therefore be considered alongside ATE. Other methods are quite stable across runs. ORB-SLAM3 and VGGT-SLAM 2.0 fail on many sequences and predict poses for only a few images (\cref{tab:pose_coverage}). DROID-SLAM, however, is accurate but slow and VRAM-intensive.

\begin{table*}[t]
\centering
\caption{\textbf{Visual odometry on Cambridge Landmarks and ETH3D SLAM.}
For each metric, Orig.\ and Low denote the original and low input
resolutions. Low resolution is $144\times256$ for Cambridge Landmarks
and $160\times256$ for ETH3D SLAM ($H\times W$).
\ours~uses low resolution in both settings.
We report success (evaluable sequences per run), FPS, latency after
sequence initialization (ms), peak VRAM (GiB), median translation
error $t$ (m), median rotation error $R$ ($^\circ$), and ATE (m).
Entries report three supplementary reruns (median across runs,
except mean peak VRAM) and may differ from the main-paper measurements.}
\label{tab:odometry-both}
\small
\setlength{\tabcolsep}{5.7pt}

\begin{tabular*}{\textwidth}{@{\extracolsep{\fill}}l rr rr rr rr rr rr rr@{}}
\toprule
& \multicolumn{2}{c}{Success$\uparrow$}
& \multicolumn{2}{c}{FPS$\uparrow$}
& \multicolumn{2}{c}{Latency$\downarrow$}
& \multicolumn{2}{c}{VRAM$\downarrow$}
& \multicolumn{2}{c}{$t\downarrow$}
& \multicolumn{2}{c}{$R\downarrow$}
& \multicolumn{2}{c}{ATE$\downarrow$} \\
\cmidrule(lr){2-3}
\cmidrule(lr){4-5}
\cmidrule(lr){6-7}
\cmidrule(lr){8-9}
\cmidrule(lr){10-11}
\cmidrule(lr){12-13}
\cmidrule(lr){14-15}
\textbf{Method}
& Orig. & Low
& Orig. & Low
& Orig. & Low
& Orig. & Low
& Orig. & Low
& Orig. & Low
& Orig. & Low \\
\midrule
\multicolumn{15}{@{}l}{\textit{Cambridge Landmarks}} \\
DROID-SLAM
& \best 8/8 & \best 8/8
& 14.8 & 14.9
& 60.8 & 60.9
& \better 5.0 & 5.0
& \better 0.32 & \best 0.33
& \best 1.00 & \best 1.14
& 5.97 & 9.17 \\
ORB-SLAM3
& 6/8 & 6/8
& \better 41.6 & \better 261.7
& \better 27.6 & \best 4.1
& n/a & n/a
& \best 0.16 & 2.80
& 6.11 & 104.38
& \best 0.70 & \better 4.52 \\
VGGT-SLAM 2.0
& \best 8/8 & 2/8
& 26.5 & \best 642.0
& 841.0 & n/a
& 5.5 & \better 4.2
& 0.55 & 21.81
& 1.67 & 74.37
& 1.36 & 7.63 \\
\addlinespace[2pt]
\ours~(ours)
& \best 8/8 & \best 8/8
& \best 66.0 & 66.0
& \best 16.7 & \better 16.7
& \best 0.6 & \best 0.6
& 0.84 & \better 0.84
& \better 1.49 & \better 1.49
& \better 1.13 & \best 1.13 \\
\midrule
\multicolumn{15}{@{}l}{\textit{ETH3D SLAM}} \\
DROID-SLAM
& \best 6/6 & \best 6/6
& 36.2 & 35.8
& 54.9 & 53.6
& \better 3.8 & \better 3.8
& \best 0.00 & \best 0.00
& \best 0.33 & \best 0.27
& \best 0.38 & \best 0.40 \\
ORB-SLAM3
& 4/6 & 3/6
& \best 221.6 & \best 762.4
& \best 5.3 & \best 2.8
& n/a & n/a
& \better 0.03 & 0.06
& \better 3.12 & 138.16
& 0.67 & 0.93 \\
VGGT-SLAM 2.0
& \best 6/6 & 3/6
& \better 213.3 & \better 650.0
& 795.8 & 804.7
& 5.4 & 5.3
& 0.07 & \better 0.05
& 6.29 & 28.05
& 0.68 & 0.94 \\
\addlinespace[2pt]
\ours~(ours)
& \best 6/6 & \best 6/6
& 94.0 & 94.0
& \better 10.7 & \better 10.7
& \best 0.7 & \best 0.7
& 0.10 & 0.10
& 5.22 & \better 5.22
& \better 0.54 & \better 0.54 \\
\bottomrule
\end{tabular*}
\end{table*}

\subsection{Number of Recurrent Iterations}\label{suppl:loops}
Since the trunk of \ours~is a single recurrent block (\cref{subsec:architecture}), the number of iterations $K$ can be chosen at inference. \Cref{fig:loop-curve} shows that accuracy is stable within the trained range $K\in[8,16]$, degrades gracefully beyond it, and drops sharply for $K\leq4$.

\subsection{Parameter Count}\label{suppl:params}
\begin{figure}[t]
\centering
\includegraphics[width=0.99\linewidth]{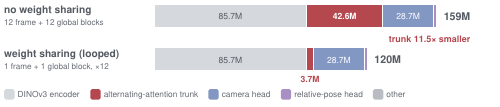}
\caption{\textbf{Parameter breakdown} without and with weight sharing (rows ``+ relative pose'' and ``+ weight sharing'' of \cref{tab:arch-ablation}). The feature projectors, used only for distillation, are not counted.}
\label{fig:param-breakdown}
\end{figure}

\Cref{fig:param-breakdown} breaks down the parameters of the last two rows of \cref{tab:arch-ablation}. Sharing the weights of one frame and one global attention block across the 12 iterations makes the trunk $11.5\times$ smaller (42.6M to 3.7M parameters), so that, without the pretrained DINOv3 encoder, \ours~has 34M parameters instead of 73M. The camera head, designed as VGGT-$\Omega$'s, then holds most of the remaining parameters (29M).

\subsection{Where the Efficiency Comes From}\label{suppl:cost-ladder}
\begin{figure}[t]
\centering
\includegraphics[width=0.99\linewidth]{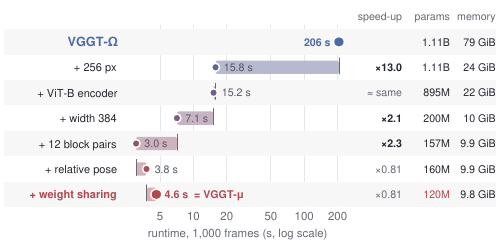}
\caption{\textbf{Cost breakdown from VGGT-$\Omega$ to \ours.} Runtime (log scale), parameters, and peak memory for 1,000 frames, applying one change per row. Rows after ``256 px'' are built in our codebase starting from VGGT-$\Omega$'s architecture. For all baselines throughout the paper, we measure runtime and peak
  memory with the depth head disabled unless it is explicitly needed.}
\label{fig:cost-ladder}

\end{figure}

\Cref{fig:cost-ladder} goes from VGGT-$\Omega_{512}$ to \ours~one change at a time, timing pose estimation for 1,000 square frames on one GH200. Lowering the resolution to 256 pixels gives the largest gain ($13\times$), followed by the narrower trunk ($2.1\times$, width 1024 to 384) and halving its depth ($2.3\times$, 24 to 12 block pairs). The smaller encoder barely matters at this length. The pairwise relative pose head and weight sharing each add about 20\% runtime: weight sharing reduces parameters (160M to 120M), not computation. The small slowdown from weight sharing comes from the per-step gates that condition the shared block on its loop position~\cite{burzio2026dejaview}, which add element-wise operations to every block.

\subsection{Computational Effect of Sequence Length}

\begin{figure}[!htb]
\centering
\includegraphics[width=0.99\linewidth]{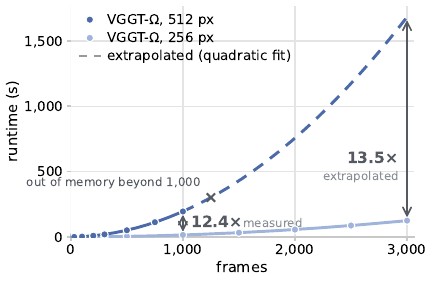}
\caption{\textbf{VGGT-$\Omega$ runtime scaling.} As the sequence length extends, the runtime is dominated by the attention mechanism, which scales as $\mathcal{O}(R^4)$ with the resolution $R$.} 
\label{fig:runtime-blowup}
\end{figure}
\paragraph{Runtime.} For $N$ frames with $P$ patches each, global attention scales as $\mathcal{O}((NP)^2)$. Since $P$ scales as $\mathcal{O}(R^2)$ with the resolution $R$, the runtime scales as $\mathcal{O}(N^2R^4)$. Thus, halving the resolution (\eg~from 512 to 256) leads, asymptotically, to a $16\times$ lower runtime (see \cref{fig:runtime-blowup}).

\begin{figure}[!htb]
\centering
\includegraphics[width=0.99\linewidth]{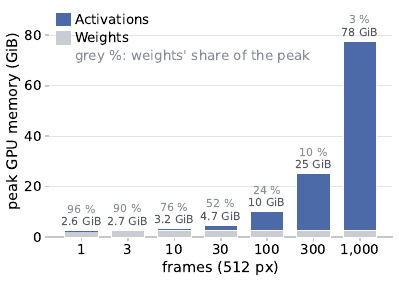}
\caption{\textbf{VGGT-$\Omega$ VRAM proportion for different sequence lengths.}  For long sequences, the attention activations dominate, whereas for short sequences the weights dominate.}
\label{fig:vram-breakdown}
\end{figure}

\paragraph{VRAM.} Using FlashAttention, VRAM scales linearly ($\mathcal{O}(NP)$) with the number of patches. Thus, for long sequences, the resolution is not as important. For short sequences, the weights dominate the GPU memory, whereas for long sequences, activations dominate (see \cref{fig:vram-breakdown}).

\begin{figure}[!htb]
\centering
\includegraphics[width=0.99\linewidth]{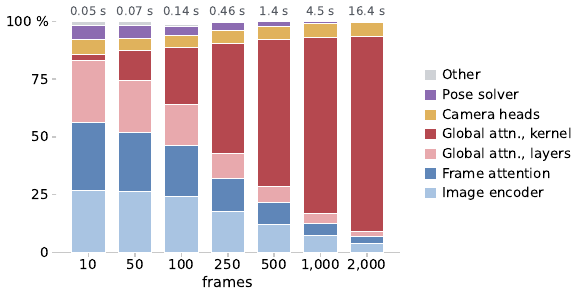}
\caption{\textbf{Runtime breakdown of \ours.} Share of the forward pass spent in each component, with the total time above each bar. For short sequences, the per-frame parts (image encoder, frame attention) dominate. Beyond a few hundred frames, the global-attention kernel takes over, while pose averaging stays a small fraction.}
\label{fig:runtime-breakdown}
\end{figure}

\paragraph{Runtime of \ours.} \Cref{fig:runtime-breakdown} breaks the runtime of \ours~down by component.

\subsection{Mixed-Scene Evaluation}\label{suppl:scene-clustering}

\Cref{tab:mixed-scene-full,tab:scene-separation-full} extend \cref{tab:mixed-scene,tab:scene-separation} in the main paper with all baselines.

\paragraph{Mixes.}
Each mix shuffles 8 frames from each of 3 distinct test scenes, taking the frames of a scene from one random sequence. Each dataset has 200 mixes drawn with a fixed seed and shared by all methods.

\paragraph{Relative pose.}
We score only same-scene pairs, using the AUC of the maximum of the rotation and translation-direction errors. We compare three protocols. In \emph{alone}, an oracle upper bound, each scene is reconstructed separately. In \emph{mixed}, all 24 frames form one input. In \emph{cluster, then reconstruct}, the frames are clustered, and each cluster is reconstructed separately. Baselines cluster with MegaLoc~\cite{berton2025megaloc} descriptors and \ours~with its own $\hat{c}_{ij}$. A same-scene pair split across clusters counts as a failure ($180^\circ$).

\paragraph{Distractors.} For the \emph{+3 distractors} setting of \cref{fig:teaser}(c), each scene of a mix is reconstructed with three single frames from other scenes of the same dataset shuffled in (drawn with a separate fixed seed, so the scenes are those of the mixes). As for \emph{alone}, only pairs within the scene are scored. Pairs involving a distractor are ignored. All three settings of \cref{fig:teaser}(c) use a single forward pass for every method, \ours~included, without clustering. \Cref{tab:mixed-drop} gives the per-dataset numbers.

\begin{table*}[t]
\centering
\caption{\textbf{Data of \cref{fig:teaser}(c).} AUC@30$^\circ$ on same-scene pairs, each in a single forward pass: every scene alone (8 frames), with 3 distractor frames from other scenes, and 3 scenes mixed (24 frames). The last column is the drop from alone to mixed in the mean.}
\label{tab:mixed-drop}
\small
\setlength{\tabcolsep}{3.5pt}
\begin{tabular}{@{}l rrr rrr rrr rrr r@{}}
\toprule
& \multicolumn{3}{c}{\textbf{RE10K}} & \multicolumn{3}{c}{\textbf{ScanNet++}} & \multicolumn{3}{c}{\textbf{MegaDepth}} & \multicolumn{3}{c}{\textbf{Mean}} & \\
\cmidrule(lr){2-4}\cmidrule(lr){5-7}\cmidrule(lr){8-10}\cmidrule(lr){11-13}
\textbf{Method} & alone & +3 & mixed & alone & +3 & mixed & alone & +3 & mixed & alone & +3 & mixed & \textbf{Drop} \\
\midrule
$\pi^3$             & 88.4 & 82.5 & 57.1 & 91.8 & 88.2 & 51.8 & 93.8 & 92.7 & 82.1 & 91.3 & 87.8 & 63.7 & $-27.6$ \\
VGGT-$\Omega_{512}$ & 90.3 & 74.0 & 56.2 & 81.3 & 70.8 & 47.8 & 93.0 & 90.2 & 83.0 & 88.2 & 78.3 & 62.3 & $-25.9$ \\
DA3-G               & 90.1 & 75.8 & 52.8 & 81.0 & 67.9 & 41.0 & 89.2 & 81.2 & 60.9 & 86.8 & 75.0 & 51.5 & $-35.2$ \\
$R^3$               & 82.5 & 66.4 & 39.9 & 72.1 & 56.5 & 28.2 & 76.9 & 68.1 & 51.4 & 77.2 & 63.7 & 39.8 & $-37.4$ \\
\dejaview           & 89.3 & 74.2 & 55.9 & 93.2 & 78.5 & 51.0 & 81.8 & 73.1 & 57.4 & 88.1 & 75.2 & 54.8 & $-33.3$ \\
VGGT                & 79.4 & 65.3 & 46.0 & 72.0 & 59.6 & 33.5 & 89.3 & 78.8 & 51.6 & 80.2 & 67.9 & 43.7 & $-36.5$ \\
\ours               & 92.0 & 91.7 & 91.2 & 86.9 & 86.0 & 83.5 & 87.8 & 87.3 & 86.9 & 88.9 & 88.3 & 87.2 & $-1.7$ \\
\bottomrule
\end{tabular}
\end{table*}

\paragraph{Scene separation.}
Retrieval methods score a pair by the cosine similarity of their L2-normalized global descriptors, clipped at 0 and computed at each method's own resolution. \ours~uses its symmetrized edge confidence $\hat{c}_{ij}$ from one pass over the mix. We cluster with average linkage on one minus this score, so that a single spurious edge cannot merge two scenes, and cut at one threshold per method. The threshold is tuned on mixes drawn with a different seed to maximize the mean ARI over the three datasets. We report the Adjusted Rand Index (ARI)~\cite{hubert1985comparing}, where 100 is a perfect partition, 0 is chance, and the Success Rate (SR), the fraction of mixes partitioned exactly.

\begin{table}[t]
\centering
\caption{\textbf{Relative pose estimation on disconnected view graphs (all baselines).}
Each input holds 8 frames from each of 3 unrelated scenes, shuffled. We report pose AUC on same-scene pairs, comparing one forward pass to first running MegaLoc~\cite{berton2025megaloc} and reconstructing each cluster separately. \ours~clusters with $\hat{c}_{ij}$.}
\label{tab:mixed-scene-full}
\small
\setlength{\tabcolsep}{3pt}
\begin{tabular}{@{}l c rr rr rr@{}}
\toprule
\textbf{Method}
& \textbf{MegaLoc}
& \multicolumn{2}{c}{\textbf{RE10K}}
& \multicolumn{2}{c}{\textbf{ScanNet++}}
& \multicolumn{2}{c}{\textbf{MegaDepth}} \\
\cmidrule(lr){3-4}
\cmidrule(lr){5-6}
\cmidrule(lr){7-8}
& 
& $3^\circ$ & $30^\circ$
& $3^\circ$ & $30^\circ$
& $3^\circ$ & $30^\circ$ \\
\midrule
\multicolumn{8}{@{}l}{\textit{Single forward pass}} \\
\dejaview            & \xmark & 14.1 & 55.9 & \good{14.6} & \good{51.0} & 10.8 & 57.4 \\
DA3-G                & \xmark & \better{19.5} & 52.8 & 10.9 & 41.0 & 20.8 & 60.9 \\
$R^3$                & \xmark & 11.2 & 39.9 &  6.3 & 28.2 &  5.5 & 51.4 \\
$\pi^3$              & \xmark & \good{15.2} & \better{57.1} & \better{19.5} & \better{51.8} & \better{40.0} & \good{82.1} \\
VGGT                 & \xmark &  7.9 & 46.0 &  4.6 & 33.5 & 10.0 & 51.6 \\
VGGT-$\Omega_{512}$  & \xmark & 13.9 & \good{56.2} &  5.7 & 47.8 & \good{33.9} & \better{83.0} \\
VGGT-$\Omega_{256}$  & \xmark & 11.9 & 55.6 &  0.8 & 37.6 &  8.5 & 57.2 \\
\ours                & \xmark & \best{52.1} & \best{91.2} & \best{42.0} & \best{83.5} & \best{49.0} & \best{86.9} \\
\midrule
\multicolumn{8}{@{}l}{\textit{Cluster, then reconstruct}} \\
\dejaview            & \cmark & 50.2 & 88.0 & \better{54.1} & \better{82.0} & 36.8 & 80.5 \\
DA3-G                & \cmark & \better{56.7} & \good{88.7} & 24.0 & 70.3 & \better{61.1} & \good{87.7} \\
$R^3$                & \cmark & 34.3 & 81.4 & 21.0 & 62.5 & 18.5 & 75.9 \\
$\pi^3$              & \cmark & 52.8 & 87.1 & \best{57.6} & \good{81.3} & \best{67.2} & \best{92.6} \\
VGGT                 & \cmark & 26.0 & 78.3 & 14.4 & 62.8 & 35.1 & 87.2 \\
VGGT-$\Omega_{512}$  & \cmark & \best{57.6} & \better{88.9} & 18.7 & 71.4 & \good{61.0} & \better{91.7} \\
VGGT-$\Omega_{256}$  & \cmark & 43.7 & 85.5 &  1.7 & 52.7 & 31.5 & 82.9 \\
\ours                & \xmark & \good{54.7} & \best{92.0} & \good{47.8} & \best{86.4} & 51.7 & \good{87.7} \\
\bottomrule
\end{tabular}
\end{table}

\begin{table}[t]
\centering
\caption{\textbf{Scene separation on mixed collections (all baselines).}
Each input holds 8 frames from each of 3 unrelated scenes, shuffled together (the mixes of \cref{tab:mixed-scene-full}). Retrieval methods cluster frames by average linkage on the cosine similarity of their global descriptors, \ours~by its predicted edge confidence $\hat{c}_{ij}$. Each method uses one threshold for all datasets, tuned on separate mixes. We report the Adjusted Rand Index (ARI) of the clustering and the Success Rate (SR), the fraction of mixes whose scenes are all recovered exactly.}
\label{tab:scene-separation-full}
\small
\setlength{\tabcolsep}{4pt}
\begin{tabular}{@{}l rr rr rr@{}}
\toprule
\textbf{Method}
& \multicolumn{2}{c}{\textbf{RE10K}}
& \multicolumn{2}{c}{\textbf{ScanNet++}}
& \multicolumn{2}{c}{\textbf{MegaDepth}} \\
\cmidrule(lr){2-3}\cmidrule(lr){4-5}\cmidrule(lr){6-7}
& ARI & SR & ARI & SR & ARI & SR \\
\midrule
NetVLAD~\cite{arandjelovic2016netvlad}   & \good{93.2} & 77.5 & 67.4 & 2.0 & 66.6 & 0.0 \\
CosPlace~\cite{Berton_CVPR_2022_CosPlace}       & 90.8 & 63.0 & 33.4 & 0.0 & 58.8 & 0.0 \\
EigenPlaces~\cite{berton2023eigenplaces} & \better{93.4} & 69.5 & 48.7 & 2.5 & 66.0 & 0.0 \\
DINOv2~\cite{oquab2023dinov2}            & 88.3 & 65.0 & 53.5 & 0.5 & \good{96.1} & \good{75.5} \\
SALAD~\cite{Izquierdo_CVPR_2024_SALAD}          & 92.7 & \good{79.5} & 60.1 & 5.5 & 72.2 & 1.5 \\
BoQ~\cite{Ali-bey_2024_CVPR}                    & 91.9 & 78.5 & \good{72.8} & \good{8.5} & 78.3 & 9.0 \\
MegaLoc~\cite{berton2025megaloc}         & 91.8 & \better{81.5} & \better{86.7} & \better{31.0} & \better{98.1} & \better{80.0} \\
\ours                                    & \best{99.5} & \best{98.5} & \best{96.4} & \best{83.0} & \best{98.6} & \best{94.0} \\
\bottomrule
\end{tabular}
\end{table}

\subsection{Training on Disconnected View Graphs}

\begin{figure}[t]
\centering
\includegraphics[width=0.99\linewidth]{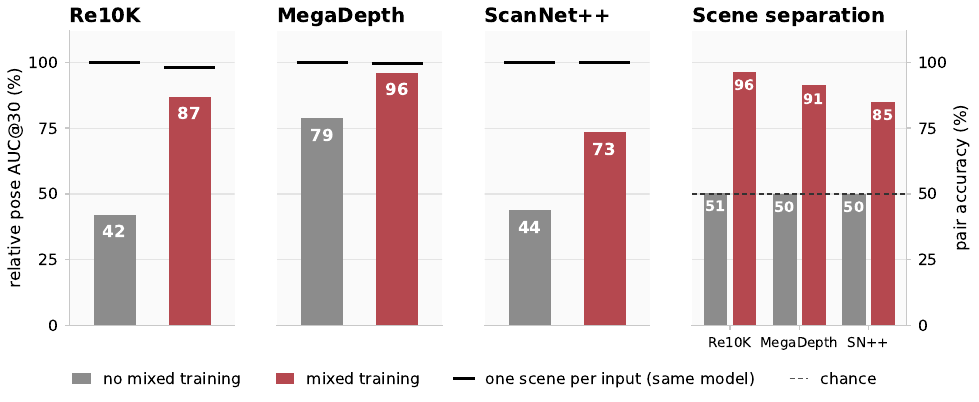}
\caption{\textbf{Mixing ablation.} Comparing training with and without sequence mixing, each trained from scratch for one day on 4 GPUs (the ablation schedule of \cref{subsec:results:ablations}), hence the lower AUC than the full model. We report the accuracy as a percentage of single-scene evaluation performance. Training on mixed sequences has almost no impact on one-scene performance (\ie, black lines are almost equally high), yet it significantly improves performance on mixed-scene sequences. Pair accuracy is the fraction of pairs correctly classified as same or different scene by thresholding $\hat{c}_{ij}$ at 0.5, averaged over same-scene and different-scene pairs.}
\label{fig:mixed-ablation}
\end{figure}

In \cref{fig:mixed-ablation}, we compare the performance on input collections that are mixed between scenes after training with and without scene mixing during training. The pose accuracy on mixed inputs improves substantially using the training protocol, while accuracy on standard inputs (one scene) remains similar. 

\subsection{Relative versus Absolute Pose Regression}\label{suppl:rel-vs-abs-pose}

Absolute pose regression assigns one pose to each camera, thereby inducing
a set of relative poses that is consistent across all paths.
Specifically, let $\mathbf{T}_i$ denote the world-to-camera pose of camera $i$.
The induced relative transform $\mathbf{T}_{ij} = \mathbf{T}_j \mathbf{T}_i^{-1}$ satisfies
\begin{equation}
     \mathbf{T}_{ij} = \mathbf{T}_{kj} \mathbf{T}_{ik}
    \qquad \text{for all distinct } i,j,k.
\end{equation}
Direct relative pose regression does not enforce this constraint:
different paths between the same cameras may yield different relative
transforms.

We investigate this distinction using the raw pairwise predictions of
VGGT-$\mu$, before pose averaging. We sample two cameras $i$ and $j$
and compare the directly predicted transform with the transform obtained
by composing an $L$-edge random walk between the same endpoints.
We measure geodesic rotation disagreement and endpoint-position
disagreement, with translation converted to meters using the
model-predicted scale. All directed edges are retained regardless of confidence.

\begin{figure}[t]
    \centering
    \includegraphics[width=0.99\linewidth]{
        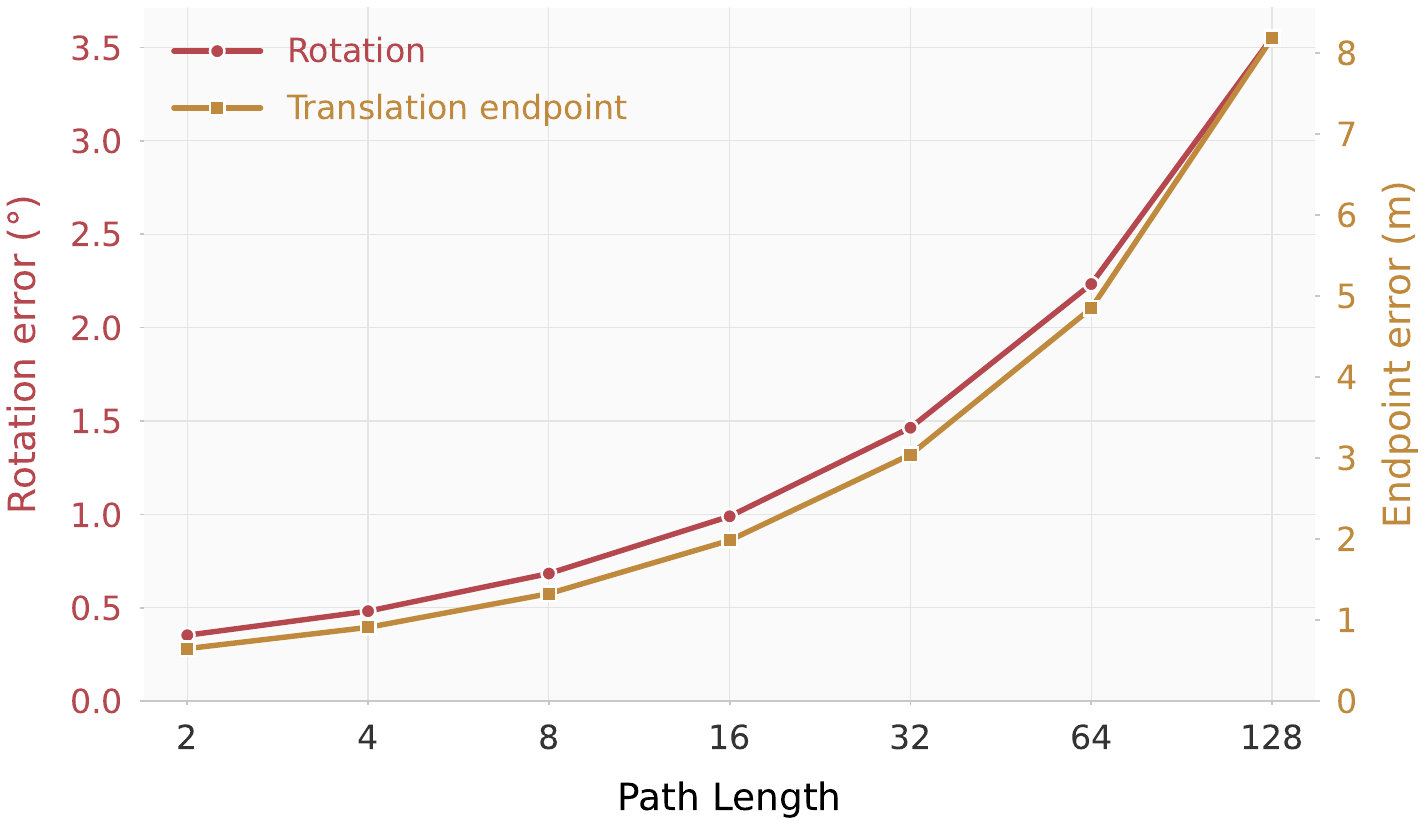
    }
    \caption{
        \textbf{Path consistency of VGGT-$\mu$ relative poses.}
        Transforms composed along random walks are compared with the
        directly predicted transform between the same endpoints.
        Curves show mean geodesic rotation disagreement
        (left axis) and mean endpoint-position disagreement
        (right axis), averaged equally across five Cambridge
        Landmarks scenes. Path length denotes the number of composed
        edges.
    }
    \label{fig:path-consistency}
\end{figure}

As shown in \cref{fig:path-consistency}, both rotation and endpoint-position
disagreement increase with path length. These results demonstrate
accumulating path inconsistency in the raw relative pose predictions.
Consequently, the predicted edges cannot, in general, be represented
exactly by a single set of absolute camera poses.

\end{document}